\pdfoutput=1
\documentclass[11pt]{article}

\usepackage[preprint]{acl}

\usepackage{times}
\usepackage{latexsym}

\usepackage[T1]{fontenc}

\usepackage[utf8]{inputenc}

\usepackage{microtype}

\usepackage{inconsolata}

\usepackage{graphicx}

\usepackage{amsmath} 
\usepackage{algorithm}
\usepackage{algpseudocode}
\usepackage{todonotes}
\usepackage{lipsum}
\usepackage{booktabs}
\usepackage{paralist} %
\usepackage{hyperref}
\hypersetup{pdftitle={Towards Probabilistic Question Answering Over Tabular Data},pdfauthor={Chen Shen, Estevam Hruschka}}
\usepackage{multirow}
\usepackage{makecell}
\usepackage{bm}
\usepackage{tabularx}

\usepackage{amssymb} %
\usepackage{xcolor} %
\usepackage{array} %
\usepackage{caption} %
\newcommand{\cmark}{\textcolor{green!70!black}{\checkmark}} %
\newcommand{\xmark}{\textcolor{red}{\ding{55}}} %

\usepackage{pifont} %
\usepackage{xspace}
\usepackage[most]{tcolorbox}
\tcbuselibrary{breakable}
\newenvironment{promptbox}[1]{%
  \begin{tcolorbox}[
    colback=gray!5!white,
    colframe=gray!50!black,
    fonttitle=\bfseries\small,
    title={#1},
    breakable,
    left=4pt, right=4pt, top=4pt, bottom=4pt,
    before upper={\setlength{\parskip}{2pt}}
  ]
  \ttfamily\footnotesize
}{%
  \end{tcolorbox}
}

\definecolor{ForestGreen}{rgb}{0.0, 0.66, 0.47}
\definecolor{RubineRed}{rgb}{1.0, 0.0, 0.31}

\newcommand{\benchmarkname}{\ensuremath{\mathsf{LUCARIO}}\xspace}

\newsavebox{\placeholderbox}

\title{Towards Probabilistic Question Answering Over Tabular Data}

\author{
  Chen Shen \\
  Megagon Labs \\
  \texttt{\detokenize{chen_s@megagon.ai}}
  \And
  Estevam Hruschka \\
  Megagon Labs \\
  \texttt{estevam@megagon.ai}
}

\begin{document}
\maketitle

\begin{abstract}
Progress in question answering (QA) over tabular data has enabled reliable factual retrieval from relational tables. However, many real-world questions are probabilistic, requiring reasoning under uncertainty and latent conditional dependencies that are not directly stored in individual cells. We introduce \benchmarkname, a large-scale benchmark for probabilistic QA over real-world tabular datasets, covering diverse domains, dependency patterns, and natural language variations. We propose Auto-BN, a fully automatic neuro-symbolic framework that induces a Bayesian Network from raw tables, translates natural-language questions into formal probabilistic queries, and performs exact inference.
Experiments across multiple LLM backbones show that Auto-BN consistently outperforms NL2SQL, retrieval-based, and premise-based baselines, while maintaining a near-zero error rate. \benchmarkname fills a relevant gap and provides a practical and scalable testbed for advancing probabilistic reasoning over structured data.
\end{abstract}

\section{Introduction}\label{sec-intro}
Tabular data is frequently used to support decision-making across enterprise, scientific, and web applications, from CRM systems and healthcare records to logistics and financial analysis. With the rapid advancement of large language models (LLMs), question answering (QA) over structured data has emerged as a critical research frontier. Recent progress in natural language to SQL (NL2SQL or Text2SQL) systems has enabled accurate retrieval of factual information from relational tables \cite{talaei2024chess,10.1145/3627673.3679216,li2024dawn}. Factual queries such as “\textit{What is the customer’s signup date?}” can now be reliably answered by mapping natural language to SQL queries that extract precise values directly from the database.

Yet, many real-world information needs extend far beyond factual retrieval. Questions such as “\textit{What is the likelihood that a customer will churn given their recent activity?}” or “\textit{How probable is a delayed shipment if the warehouse is in Region A?}” require reasoning under uncertainty. The correct answers to these probabilistic queries involve incomplete evidence, latent and conditional dependencies that are usually not directly encoded in a single table cell, neither fully available in LLMs' parametric knowledge. Thus, current SQL-based and LLM-only approaches fall short in this setting, often hallucinating answers \cite{huang2025survey} or generalizing poorly when tasked with probabilistic reasoning \cite{chakraborty2025hallucination}. The answers to these questions are estimated from the records of a specific table, so the same churn question can have a different answer for a different CRM table.

Prior work \cite{jin2023cladder,schrader-etal-2024-quite,nafar2025reasoning} has explored probabilistic graphical models and causal inference frameworks for addressing uncertainty and probabilistic reasoning jointly with LLMs. These approaches are constrained by scalability and practicality: they typically operate only on small or synthetic datasets and rely on expert-curated structures for the underlying models. In contrast, real-world tabular data may span hundreds of attributes and tens of thousands of rows, rendering manual construction of probabilistic models infeasible. Compounding this challenge is the absence of benchmarks for probabilistic QA over large-scale tabular data. Existing benchmarks overwhelmingly target factual QA, leaving no reliable way to evaluate, compare, or systematically advance systems designed for probabilistic reasoning.

In this paper, we introduce \benchmarkname, a large-scale benchmark and a framework for probabilistic QA over tabular data. Our main contributions are:

1. A new benchmark with 10 real-world datasets, designed to capture diverse probabilistic reasoning patterns beyond factual retrieval and realistic scale.

2. Auto-BN, a fully automatic neuro-symbolic framework that (i)~induces observational Bayesian Networks (BNs) directly from large relational tables, 
(ii)~robustly translates natural language into formal probabilistic queries, 
and (iii)~performs exact probabilistic inference equipped with a multi-level semantic relaxation.

3. Extensive empirical analysis and evaluation across diverse baselines and modeling paradigms, demonstrating consistent improvements and providing insights into the challenges and opportunities of probabilistic QA over large-scale tabular data.

Our results show that the proposed hybrid approach not only improves accuracy on complex probabilistic questions but also enhances interpretability and robustness. Beyond the proposed system, \benchmarkname fills a key gap in existing resources by providing a realistic and scalable testbed for evaluating probabilistic question answering over tabular data, supporting systematic research on methods that must reason under uncertainty.

\section{Related Work}\label{sec-prelim}

\noindent \textbf{Question Answering over Tabular Data: }
Question answering (QA) over tabular data has become an increasingly active research area, with models such as TAPAS \cite{herzig2020tapas}, TURL \cite{deng2022turl}, and Chase-SQL \cite{pourreza2024chase} demonstrating strong performance on factual queries by translating natural language into structured queries like SQL. These systems excel when answers can be directly retrieved from table cells, e.g., \textit{“What is the customer’s signup date?”}. 
Nonetheless, they struggle with queries requiring reasoning beyond direct retrieval—particularly those involving uncertainty or probabilistic dependencies. 

Recent systems have incorporated more complex reasoning pipelines. TQAgent \cite{zhao2025tqagent} integrates knowledge graphs and tree-structured reasoning paths to handle multi-step queries across diverse table schemas. TableBench \cite{wu2025tablebench} spans fact checking, numerical reasoning, data analysis, and visualization, and TReB \cite{li2026treb} defines 26 table-reasoning sub-tasks; neither targets conditional-probability questions answered from a joint distribution induced from the table. While such efforts extend the scope of tabular QA, they remain ill-suited for probabilistic reasoning tasks that demand modeling latent dependencies and conditional relationships not explicitly represented in the data.

\noindent \textbf{Benchmarks for Probabilistic Dependency Reasoning:}
Evaluating how language models reason about uncertainty and dependencies among variables has motivated the development of several recent benchmarks. 

Some benchmarks explicitly focus on causal reasoning. \textbf{CausalQA} \cite{bondarenko2022causalqa} and \textbf{QRData} \cite{liu2024llms} evaluate models’ ability to identify causal structures and reason about interventions and counterfactuals. These datasets target reasoning tasks grounded in causal graphical models and structural assumptions about data-generating processes.

Other benchmarks emphasize probabilistic reasoning more broadly, often relying on Bayesian networks or textual descriptions of probabilistic relationships. \textbf{BLInD} \cite{nafar2025reasoning} provides natural language queries grounded in predefined Bayesian Networks (BNs), requiring models to reason over structured uncertainty and produce numeric probability estimates. \textbf{CLADDER} \cite{jin2023cladder} evaluates reasoning across different levels of dependency queries, including associational and causal questions derived from symbolic graphs. \textbf{QUITE} \cite{schrader-etal-2024-quite} tests Bayesian reasoning using textual descriptions of probabilistic relationships, assessing models’ ability to quantify uncertainty from structured descriptions.

While impactful, these benchmarks share notable limitations. They typically rely on small, predefined Bayesian Networks (e.g., QUITE networks average only 6 nodes and 3 states per node, yielding fewer than 50 unique one-hop reasoning paths on average). This restricts both the diversity of reasoning patterns and the difficulty of inference. 
Moreover, they are often domain-specific (e.g., healthcare) and require significant human curation, hindering scalability and generalization.

\noindent \textbf{Positioning Our Work:}

\begin{figure}[ht]
  \centering
  \includegraphics[width=0.45\textwidth]{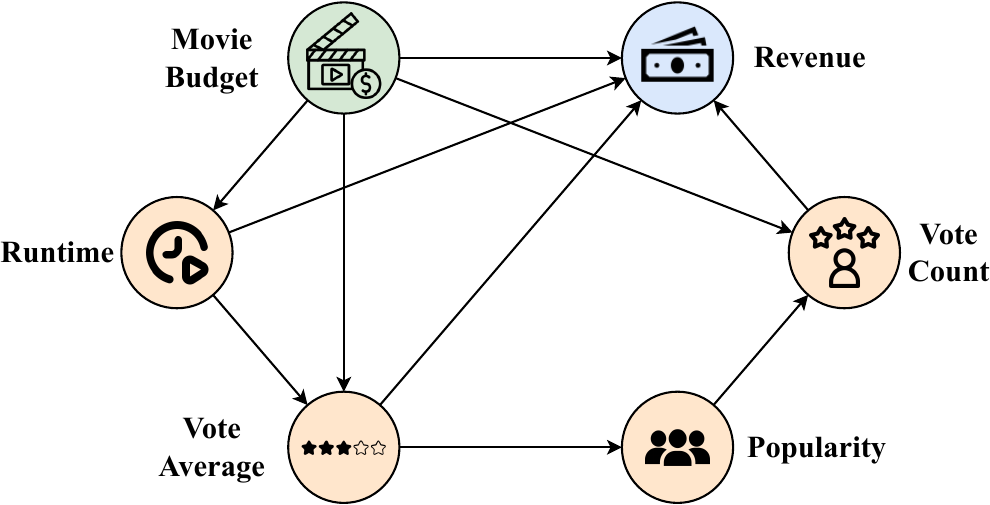}
  \caption{Expert Verified Data-Driven Bayesian Network Structure on Movie table}
  \label{fig:causal_graph_example}
\end{figure}

Distinct from prior work, \textbf{\benchmarkname} is constructed through a scalable, data-driven and human-validated process that enables probabilistic reasoning directly over tabular data. From real-world relational tables spanning diverse domains, Bayesian Networks are semi-automatically induced (not requiring manual structure design, while incorporating human knowledge and validation to ensure reliability and directional dependency). Each table is thus paired with its corresponding probabilistic graph. \textbf{\benchmarkname}'s  scale allows the benchmark to capture a broader range of dependency patterns, including simple conditional relationships, multi-variable interactions, and more complex dependency structures that rarely appear in small manually designed graphs.
(Figure~\ref{fig:causal_graph_example} shows a snippet of a graph structure).

\textbf{\benchmarkname} also provides premise sets describing inter-variable dependencies, and insights (reduced set of relevant premises -- see \S~\ref{sec-benchmark}). Averaging 79k premises per table, these sets vastly exceed those in other benchmarks, offering a richer foundation for symbolic and neural reasoning under realistic, large-scale conditions where probabilistic and directional relationships arise organically from data.

Table~\ref{tab:benchmark-comparison} highlights the key differences between existing benchmarks and \benchmarkname. To our knowledge, \benchmarkname is the first benchmark to pair expert-validated Bayesian Networks, constructed semi-automatically from real, domain-diverse tables, with large-scale probabilistic QA evaluation.

\begin{table}[t]
\centering
\footnotesize %
\renewcommand{\arraystretch}{1.15}
\setlength{\tabcolsep}{3pt} %

\begin{tabular}{@{} l c c c c @{}}
\toprule
 & \textbf{LUCARIO} & \textbf{QUITE} & \textbf{CLADDER} & \textbf{BLInD} \\
\midrule
Categorical vars.   & \cmark & \cmark & \xmark & \xmark \\
Diverse domains     & \cmark & \cmark & \cmark & \xmark \\ %
Diverse Q-types     & \cmark & \xmark & \xmark & \xmark \\
Tabular data     & \cmark & \xmark & \xmark & \xmark \\ %
Scalable BNs        & \cmark & \xmark & \xmark & \xmark \\
Prob. insights      & \cmark & \xmark & \xmark & \xmark \\
\bottomrule
\end{tabular}

\caption{Comparison of \textbf{\benchmarkname} with recent benchmarks for probabilistic or causal reasoning.}
\label{tab:benchmark-comparison}
\end{table}

\section{\benchmarkname Benchmark}\label{sec-benchmark}

We introduce \benchmarkname, a benchmark built upon the BIRD dataset \citep{li2024can}, designed to evaluate probabilistic question answering (QA) over tabular data. While BIRD focuses on translating factual queries into SQL, our benchmark introduces probabilistic queries requiring reasoning under uncertainty. The benchmark comprises three primary components: relational tables, probabilistic queries, and ground-truth answers, supplemented by auxiliary reasoning artifacts (Bayesian Networks, premises, and insights).

\subsection{Data Sources \& Benchmark Artifacts}
\label{subsec:artifacts}

\benchmarkname contains 10 real-world, large-scale relational databases from diverse domains (e.g., finance, healthcare, sports) sourced from BIRD and selected for meaningful probabilistic dependencies, feature richness, and statistical sufficiency (Appendix~\ref{subsec:data-source}). To support downstream neuro-symbolic reasoning, we pair each table with three auxiliary artifacts:

\noindent \textbf{Bayesian Network (BN)}:
A graphical model representing the directional statistical dependencies \citep{pearl1988} among the table's columns. A BN factorizes their joint distribution into local conditional probabilities, enabling exact computation of $P(X{=}x \mid \mathbf{C}{=}\mathbf{c})$ for any target and evidence.

\noindent \textbf{Premises}:
Premises are a set of textual statements characterizing the statistical relationships inherent in the tabular data. These statements function as probabilistic rules, serving as a foundation for rule-based reasoning and explanation \citep{schrader-etal-2024-quite}. Formally, a single premise corresponds to a one-hop reasoning step that represents the conditional distribution of a target node given the specific states of its parent nodes.

\noindent \textbf{Insights}:
A primary challenge of using premises as the core mechanism for the probabilistic reasoning, prediction, and explanation is the size of the premise set. Because it exhaustively covers all possible combinations of attribute states, the premise space grows exponentially for tables with ten or more attributes, introducing significant computational latency and complexity. To mitigate this, \benchmarkname includes a set of insights---a compact subset of premises identified as having the most substantial impact on the data's behavior. We define the \textit{Insight Saliency} of a variable $X$ under a conditioning context $C$ (the observed states of other columns, e.g., a customer's recent activity in the churn question of \S\ref{sec-intro}) as the Kullback--Leibler (KL) divergence between the conditional distribution $P_{X|C}$ and the base marginal distribution $P_X$:
\begin{multline}
    \text{Saliency}(X, C) = D_{\text{KL}}(P_{X|C} \parallel P_X) \\
    = \sum_{x \in X} P_{X|C}(x) \log \frac{P_{X|C}(x)}{P_X(x)}
    \label{eq:saliency}
\end{multline}
By systematically filtering for contexts $C$ that exceed a minimum saliency threshold and introduce new probability peaks, we condense the massive premise space into the most statistically significant conditional shifts, enabling efficient and highly calibrated reasoning.

\subsection{Diversity in Queries and Questions}
\label{subsec:query_diversity}

A key differentiator of \benchmarkname{} compared to prior probabilistic reasoning benchmarks is the structural and linguistic diversity of its evaluation set. Because our semi-automatic construction pipeline (\S\ref{subsec:dataset-creation}) is specifically engineered to generate these complex structures, we first define the target reasoning patterns and natural language variations that the benchmark encompasses.

\begin{figure}[ht]
   \centering
    \includegraphics[width=\columnwidth]{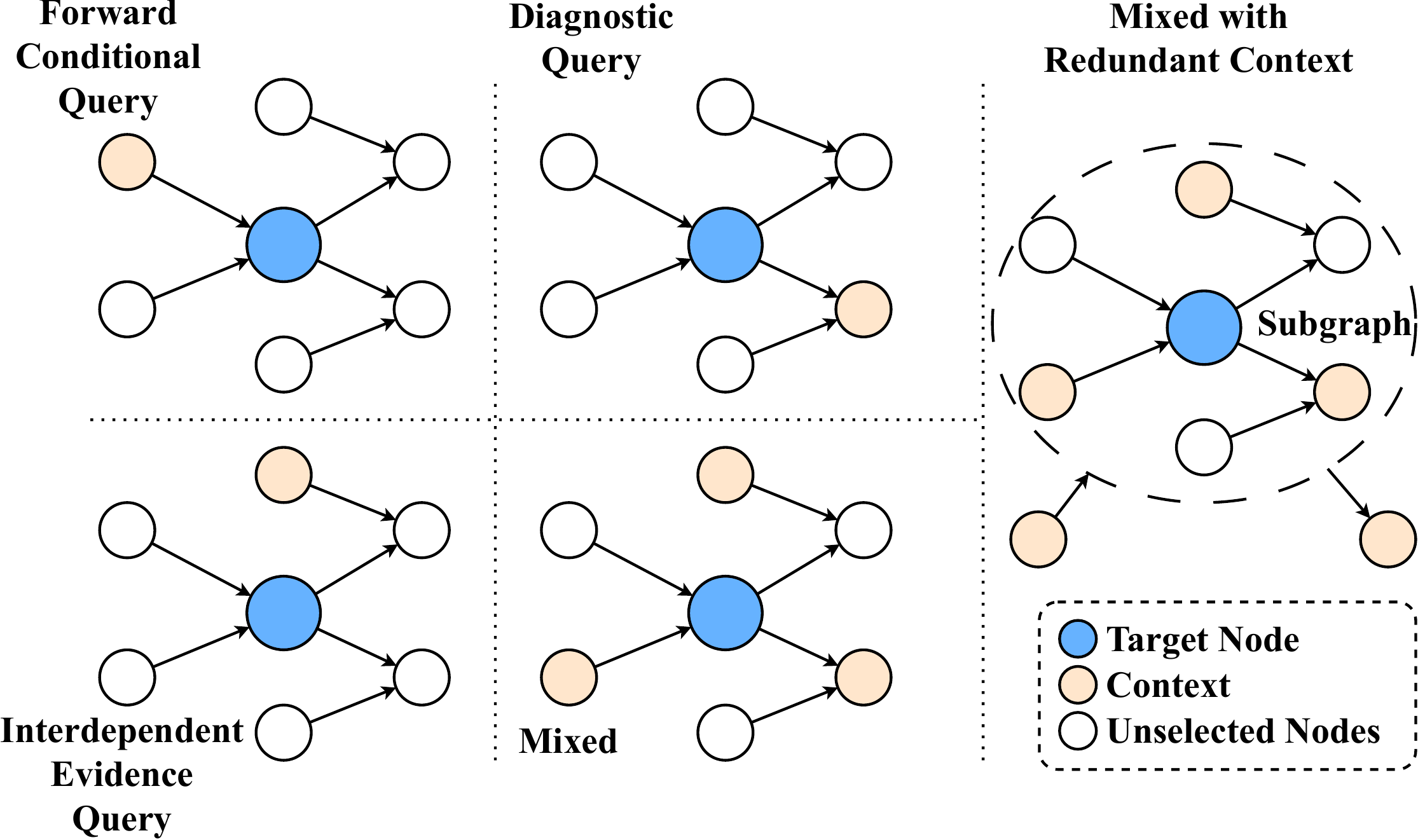}
   \caption{Probabilistic Query Types}
    \label{fig:reasoning_types}
\end{figure}

\paragraph{Probabilistic Query Types (Reasoning Patterns).}
To enable systematic evaluation of probabilistic QA, \benchmarkname{} includes a diverse set of probabilistic queries (Figure~\ref{fig:reasoning_types}) covering distinct patterns of conditional dependence among table attributes:
(i) \textbf{Forward Conditional Queries}: Probability of an outcome given observed conditions (e.g., the probability of high sales given that a marketing campaign was launched). 
(ii) \textbf{Diagnostic Queries}: Inferring the likelihood of an explanatory variable given observed evidence (e.g., the probability of a disease given a set of symptoms).
(iii) \textbf{Interdependent Evidence Queries}: Multiple correlated variables provide alternative explanations for the same observation, such that evidence supporting one explanation reduces the probability of competing ones.
(iv) \textbf{Multi-Evidence Aggregation Queries}: Handling scenarios with multiple pieces of evidence that may be overlapping or provide corroborative information.

\paragraph{Linguistic Question Variations.}
Real-world questions involving uncertainty are often expressed in diverse and nuanced ways. While many existing benchmarks for probabilistic reasoning primarily use formulations that closely mirror the symbolic representation of variables and states (e.g., a question contains ``\textit{temperature high}'' when the graph node \textit{temperature} includes a ``\textit{high}'' state), real-world user queries frequently vary in specificity, granularity, and linguistic style. 

\begin{figure}[ht]
    \centering
    \includegraphics[width=\columnwidth]{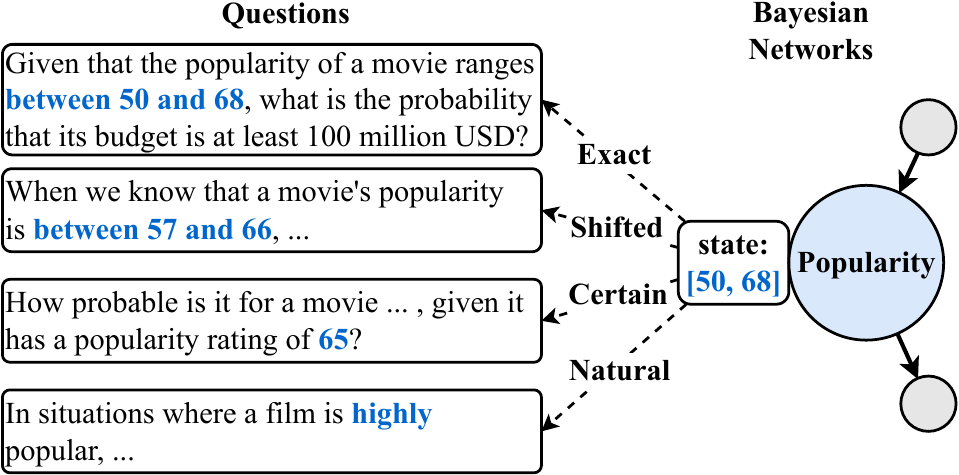}
    \caption{Question Types in \benchmarkname}
    \label{fig:question_types}
\end{figure}

To capture this variability and evaluate models' robustness under natural variation, \benchmarkname{} expands each probabilistic query into four distinct linguistic expressions (Figure~\ref{fig:question_types}):
(i) \textbf{Exact}: The question expresses the exact numeric range or symbolic state as defined in the BN.
(ii) \textbf{Shifted}: The question refers to an alternative, valid numeric range.
(iii) \textbf{Certain}: The condition is stated as a single numeric value instead of a range.
(iv) \textbf{Natural}: The condition is described using free-form natural language paraphrasing.

Table~\ref{tab:examples} gives one example for each query type.

\begin{table*}[t]
\centering
\small
\setlength{\tabcolsep}{4pt}
\begin{tabular}{>{\raggedright\arraybackslash}p{2.2cm} >{\raggedright\arraybackslash}p{10.3cm} l r}
\toprule
\textbf{Domain} & \textbf{Question} & \textbf{Variant} & \textbf{Gold} \\
\midrule
movies & \textit{Given that a film has a moderately short duration and receives an above-average rating, how likely is it to generate low earnings?} & Natural & 0.375 \\
sales in weather & \textit{When the average temperature is between 70 and 79\,\textdegree F and the dew point is at least 66\,\textdegree F, what is the probability that the calculated cooling degree is between 8 and 13
degree-days?} & Exact & 0.567 \\
hockey scoring & \textit{Given that the number of goals scored is between 4 and 7, what is the probability that the shots on goal fall between 46 and 78?} & Shifted & 0.535 \\
cookbook nutrition & \textit{Given that a recipe has between 0\% to 5\% of its calories coming from fat, 15\% to 28.6\% from protein, and 50\% to 70\% from carbohydrates, what is the probability that the cholesterol
content in the recipe is between 0 to 6.53 milligrams?} & Exact & 0.487 \\
twitter & \textit{Given that a tweet has received no likes, the sentiment analysis of the tweet is neutral, and it was posted on a Tuesday, what is the probability that the tweet was posted between 5 and 8 in the morning?}
& Certain & 0.266 \\
\bottomrule
\end{tabular}
\caption{Example \benchmarkname{} questions, one per probabilistic query type, from five domains and four linguistic variants.}
\label{tab:examples}
\end{table*}

\subsection{Dataset Creation}
\label{subsec:dataset-creation}

\benchmarkname{} is constructed via a three-stage semi-automatic pipeline incorporating expert knowledge and validation.

\noindent\textbf{Stage 1: BN Construction.}
For each selected table, continuous numerical columns are first discretized into semantically meaningful bins (e.g., mapping a raw \textit{timestamp} to \textit{morning} time range). We then apply the PC algorithm \cite{kalisch2007estimating} to induce an initial Directed Acyclic Graph (DAG) representing the probabilistic dependencies between columns. To align the benchmark with domain logic and filter spurious statistical correlations, two domain experts manually review and refine these DAGs (adding, removing, or reversing edges), injecting semantic knowledge beyond what is observable in the raw data. When they disagreed or were uncertain about an edge, they consulted conditional-independence tests and the underlying records and resolved the issue through discussion. We subsequently fit Conditional Probability Distributions (CPDs) to these validated DAGs using BDeu scoring over the table rows. 

Following established protocols in probabilistic benchmarking \citep{jin2023cladder, schrader-etal-2024-quite}, the resulting expert-validated BN serves as the ground-truth structure, from which we automatically derive the full set of premises and a distilled set of high-impact insights.

\noindent\textbf{Stage 2: Question Generation.}
Using the ground-truth BNs, we employ stratified sampling to select target variables in each BN. We traverse the graph to identify parent, child, and Markov-Blanket structures that align with our five defined probabilistic query reasoning patterns (Figure~\ref{fig:reasoning_types}). We then prompt an LLM with the table descriptions, columns statistics, BN schema, selected anchor/context nodes, and the target reasoning pattern to generate candidate natural language questions. 

To ensure high linguistic quality and logical coherence, every candidate question undergoes manual screening for fluency, correctness, and contextual relevance; sub-optimal formulations are discarded. Finally, each accepted question is expanded into four distinct linguistic variants (Figure~\ref{fig:question_types})—to evaluate model robustness against surface-level phrasing variations.

\noindent\textbf{Stage 3: Answer Computation.}
To establish the ground-truth answers, we compute exact probabilities by performing inference directly on the expert-validated BN. For a given target variable $X$ and conditioning context $\mathbf{C}$, the ground-truth probability is formally computed as $P(X{=}x \mid \mathbf{C}{=}\mathbf{c})$. Exact inference makes each answer analytically correct with respect to the expert-validated BN and its fitted CPDs, grounded in the table's data distribution, and fully reproducible.

\subsection{Dataset Statistics}

\benchmarkname includes 10 datasets (tables), each with 157 probabilistic queries. Each query is expressed in four linguistic question variants, yielding 1,570 unique probabilistic scenarios and 6,280 question–answer pairs. The statistics are shown in Table ~\ref{tab:bn-stats}, with more details in Appendix~\ref{sec-statistics}.

\begin{table}[h]
\centering
\small
\begin{tabular}{l r}
\toprule
\textbf{Statistic} & \textbf{Mean Value} \\
\midrule
Number of Nodes per BN & 12.50 \\
Number of Edges per BN & 32.60 \\
States per Node & 4.95 \\
Total Degree (avg) & 5.02 \\
\midrule
Number of Premises per BN & 79,161.8 \\
Number of Insights per BN & 2,411.8 \\
Premises per Node & 6,043.9 \\
Insights per Node & 192.8 \\
\bottomrule
\end{tabular}
\caption{Summary statistics of Bayesian Networks and probabilistic artifacts across benchmark tables.}
\label{tab:bn-stats}
\end{table}

\subsection{Benchmark Quality Evaluation}
To ensure the reliability of \benchmarkname, we conducted both structural and linguistic evaluations of the benchmark components.

\noindent \textbf{Graph Structure Evaluation.}
To evaluate the finalized benchmark artifacts, three researchers independent of the benchmark curation assessed the plausibility of a sampled subset of directional edges from the Ground-truth BNs. Provided with the graphs and underlying tabular data, the experts judged all sampled edges to be valid. Specifically, 84\% of the edges were evaluated as intuitively consistent with domain knowledge, while the remaining 16\% represented more subtle correlations that required inspection of the underlying data distributions to confirm.

\noindent \textbf{Linguistic Quality.}
Following the protocol of \citet{schrader-etal-2024-quite}, we evaluated the fluency of a random sample of 200 items (10 premises and 10 questions from each of the 10 tables) using Grammarly.\footnote{\url{https://www.grammarly.com}} The corpus achieved an overall score of 96, significantly exceeding prior benchmarks (QUITE: 82--85, CLADDER: 45, BLInD: 34).

\subsection{Evaluation Metrics}

Given a natural language question on a table and its ground-truth probability, we evaluate model performance using the following metrics:

\noindent \textbf{Error Rate:} The percentage of predictions either outside the valid probability range $[0.0, 1.0]$ or missing entirely. A fallback value of $1.0/\text{num}_{\text{states}}$ is assigned, assuming a uniform prior. This generalizes the 0.5 fallback commonly used in existing binary-state benchmarks.

\noindent \textbf{Mean Absolute Error (MAE)}: The average absolute difference between predicted and true probabilities, measuring overall estimation accuracy.

\noindent \textbf{Root Mean Squared Error (RMSE)}: The square root of the average squared error, which penalizes large deviations more heavily than MAE.

\noindent \boldmath$\textbf{Accuracy}_{x}$\unboldmath: The proportion of predictions within $\pm x$ of the ground-truth. We report $\text{Acc}_{0.02}$ and $\text{Acc}_{0.05}$ as thresholds for near-correct predictions.

\section{Auto-BN: Probabilistic Inference over Tabular Data}\label{sec-method}

While LLMs excel at factual tabular retrieval, they struggle with complex queries requiring reasoning under uncertainty. To bridge this gap, we propose \textbf{Auto-BN}, a fully automatic, end-to-end pipeline that induces BNs directly from raw tabular data and uses an LLM to translate natural language questions into exact probabilistic inference. Let $\mathcal{T}$ be a relational table and $q$ a natural language question. The goal is to compute the grounded probability $r$:
\begin{equation}
    r = P\!\left(\hat{n} = \hat{s} \;\middle|\; C\right) = \frac{\sum_{\mathbf{u}} P(\hat{n}{=}\hat{s},\, C,\, \mathbf{u})}{\sum_{\hat{s}'}\sum_{\mathbf{u}} P(\hat{n}{=}\hat{s}',\, C,\, \mathbf{u})}
    \label{eq:prob_inference}
\end{equation}
where $\hat{n}$ is the target attribute, $\hat{s}$ is its queried state, $C$ is the conditioning context derived from $q$, and $\mathbf{u}$ represents the unobserved variables. 

Auto-BN decomposes this challenge into a three-stage neuro-symbolic pipeline:
\begin{center}
$\mathcal{T} \xrightarrow{\S\ref{subsec:bnconstruct}} \mathcal{B} \xrightarrow{\S\ref{subsec:robust_translation}} q' \xrightarrow{\S\ref{subsec:multi_level_relaxation}} r$
\end{center}
where $\mathcal{B}$ is the automatically induced Bayesian Network, and $q' = (\hat{n},\, \hat{s},\, C)$ is the structured, executable query parsed from natural language. 

\subsection{Automatic BN Construction (Offline)}\label{subsec:bnconstruct}

In contrast to the expert-involved Gold BNs used for benchmark evaluation, Auto-BN induces a data-driven BN $\mathcal{B}$ from the raw table $\mathcal{T}$ via a fully automated, four-step offline pipeline: (i)~LLM-assisted column classification (binary, categorical, or numerical) based on value distributions, data types, and schema descriptions; (ii)~K-means discretization of continuous columns, utilizing an LLM to assign semantically meaningful boundary labels; (iii)~directed dependency graph induction via the PC algorithm, with algorithmic post-processing to enforce acyclicity and prune low-confidence edges; and (iv)~Conditional Probability Table (CPT) fitting via Bayesian estimation directly over the tabular records. Discretization lets parameter learning and exact inference use finite-state CPTs without the distributional assumptions (e.g., linear-Gaussian) required by continuous or hybrid BNs. It also lets one BN handle categorical and numeric columns uniformly.
Executed once per table, this pipeline yields a purely data-driven probabilistic model. By decoupling the induced graph from the human-verified ground-truth BNs, we ensure that subsequent evaluations rely exclusively on original data, thereby precluding any ground-truth leakage.
\subsection{Robust Query Translation}
\label{subsec:robust_translation}

At inference time, an LLM parses $q$ into a structured node-mapping query $q' = \langle \hat{n}, \hat{s}, C \rangle$. Drawing inspiration from execution-guided Text-to-SQL pipelines \citep{pourreza2024chase, talaei2024chess}, Auto-BN mitigates single-pass schema-linking errors via a three-phase fault-tolerant mechanism:

\noindent \textbf{Multi-Path Cross-Consistency Voting.} Auto-BN concurrently samples $N$ (default: 3) diverse parse candidates. Each is validated against the BN schema to ensure all nodes and states strictly exist in BN $\mathcal{B}$. The plurality tuple among the \textit{valid} candidates is selected, surfacing semantic ambiguity as structured uncertainty rather than a silent failure.

\noindent \textbf{Execution-Guided Iterative Refinement.} If all $N$ candidates fail validation, Auto-BN extracts the specific error trace (e.g., invalid target node) from a fresh, zero-temperature parse. This execution feedback is appended to the prompt, allowing the LLM to self-correct its schema linking for up to $K_{\text{corr}}$ rounds (default: $K_{\text{corr}}=2$), recovering the vast majority of initial vocabulary mismatches.

\noindent \textbf{Soft Fallback and Typed Error Taxonomy.} Upon depleting the correction budget, Auto-BN assigns a typed error (e.g., \texttt{invalid\_state}). If the initial voting phase produced any parseable candidate, this best-effort mapping is forwarded to the inference fallback chain (\S\ref{subsec:multi_level_relaxation}) instead of being discarded, maximizing structural recovery before abstention. Together, these mechanisms reduce the parser error rate from 5.6--20.8\% under single-pass parsing to near zero across three LLM backbones (Appendix~\ref{subsec:error_analysis}, Table~\ref{tab:ablation_robust}). 

\subsection{Inference with Multi-Level Relaxation}
\label{subsec:multi_level_relaxation}

A structurally valid query $q'$ is executed via variable elimination (Eq.~\ref{eq:prob_inference}). To handle queries suffering from persistent vocabulary mismatches or execution-time logical exceptions (including those forwarded via the soft fallback mechanism), Auto-BN applies a four-level heuristic relaxation chain: (i) \textit{Semantic Node Relaxation} replaces an unmapped $\hat{n}$ with the closest valid BN node via cosine similarity using sentence embeddings; 
(ii) \textit{Semantic State Relaxation} similarly maps invalid states $\hat{s}$ or states within $C$ to their nearest valid neighbors; 
(iii) \textit{Context Pruning} iteratively strips conflicting evidence variables from $C$; and (iv) \textit{Marginal Fallback} returns the unconditioned prior $P(\hat{n}=\hat{s})$ if all conditioned variants fail. Appendix~\ref{appendix:autobn_tool} describes Auto-BN-Tool, an agentic tool-calling extension of this pipeline, and Appendix~\ref{appendix:prompts} provides the full prompt templates.

\section{Experimental Evaluation}\label{sec-exp}

We evaluate the Auto-BN framework and baselines on the 10 datasets in \benchmarkname benchmark.

\begin{table*}[!ht]
\centering
\small
\setlength{\tabcolsep}{2pt}
\renewcommand{\arraystretch}{0.8}
\resizebox{\linewidth}{!}{
\begin{tabular}{l l l c c c c c}
\toprule
\textbf{Type} & \textbf{Method} & \textbf{LLM} & \textbf{\% Acc\textsubscript{0.02}} $\uparrow$ & \textbf{\% Acc\textsubscript{0.05}} $\uparrow$ & \textbf{MAE} $\downarrow$ & \textbf{RMSE} $\downarrow$ & \textbf{\% Error Rate} $\downarrow$ \\
\specialrule{1.2pt}{1pt}{3pt}

\multirow{1}{*}{\rotatebox[origin=c]{90}{\textbf{}}} 
  & Random & - & $4.5 \pm 1.8$ & $11.1 \pm 2.7$ & $0.362 \pm 0.021$ & $0.447 \pm 0.023$ & $0.0 \pm 0.0$ \\

\specialrule{1.2pt}{1pt}{3pt}

\multirow{11}{*}{\rotatebox[origin=c]{90}{\textbf{Factual}}} 
  & \multirow{3}{*}{LLM + Table} & GPT-4o & $11.1 \pm 6.8$ & $24.4 \pm 9.6$ & $0.171 \pm 0.046$ & $0.237 \pm 0.060$ & $0.0 \pm 0.0$ \\
  & & Llama-3.3-70B & $11.1 \pm 5.1$ & $25.4 \pm 8.1$ & $0.203 \pm 0.041$ & $0.284 \pm 0.052$ & $0.0 \pm 0.0$ \\
  & & Mixtral-8x22B & $10.8 \pm 4.3$ & $23.4 \pm 8.2$ & $0.179 \pm 0.042$ & $0.243 \pm 0.051$ & $0.0 \pm 0.0$ \\
  \cmidrule{2-8}
  & \multirow{2}{*}{Reasoning Models} & GPT-5.4 & $11.4 \pm 4.6$ & $25.3 \pm 7.9$ & $0.199 \pm 0.067$ & $0.278 \pm 0.074$ & $0.0 \pm 0.0$ \\
  & & DeepSeek-R1 & $11.4 \pm 6.9$ & $21.9 \pm 12.8$ & $0.204 \pm 0.051$ & $0.276 \pm 0.064$ & $0.0 \pm 0.0$ \\
  \cmidrule{2-8}
    & \multirow{3}{*}{TableRAG} & GPT-4o & $7.6 \pm 5.0$ & $18.6 \pm 7.6$ & $0.220 \pm 0.037$ & $0.297 \pm 0.046$ & $0.0 \pm 0.0$ \\
  & & Llama-3.3-70B & $7.4 \pm 4.9$ & $19.4 \pm 7.3$ & $0.226 \pm 0.052$ & $0.303 \pm 0.062$ & $0.0 \pm 0.0$ \\
  & & Mixtral-8x22B & $6.6 \pm 4.1$ & $18.7 \pm 8.1$ & $0.217 \pm 0.046$ & $0.285 \pm 0.052$ & $0.0 \pm 0.0$ \\
  \cmidrule{2-8}
  & \multirow{3}{*}{NL2SQL} & GPT-4o & $13.2 \pm 9.7$ & $25.1 \pm 14.5$ & $0.190 \pm 0.062$ & $0.260 \pm 0.075$ & $6.2 \pm 7.9$ \\
  & & Llama-3.3-70B & $14.6 \pm 9.9$ & $25.8 \pm 13.8$ & $0.201 \pm 0.067$ & $0.278 \pm 0.082$ & $3.2 \pm 6.2$ \\
  & & Mixtral-8x22B & $7.3 \pm 5.7$ & $17.3 \pm 7.4$ & $0.186 \pm 0.044$ & $0.253 \pm 0.063$ & $52.8 \pm 30.3$ \\

\specialrule{1.2pt}{1pt}{3pt}

\multirow{9}{*}{\rotatebox[origin=c]{90}{\textbf{Prob. Premises}}} 
  & \multirow{3}{*}{LLM+Premise(Vector)} & GPT-4o & $8.4 \pm 4.4$ & $20.5 \pm 6.3$ & $0.178 \pm 0.047$ & $0.243 \pm 0.062$ & $0.0 \pm 0.0$ \\
  & & Llama-3.3-70B & $6.7 \pm 4.8$ & $16.8 \pm 7.4$ & $0.227 \pm 0.050$ & $0.287 \pm 0.056$ & $0.1 \pm 0.4$ \\
  & & Mixtral-8x22B & $4.8 \pm 2.8$ & $14.2 \pm 4.7$ & $0.299 \pm 0.044$ & $0.404 \pm 0.048$ & $2.6 \pm 2.3$ \\
  \cmidrule{2-8}
  & \multirow{3}{*}{LLM+Premise(BM25)} & GPT-4o & $8.7 \pm 5.8$ & $19.3 \pm 8.0$ & $0.184 \pm 0.029$ & $0.253 \pm 0.036$ & $0.0 \pm 0.0$ \\
  & & Llama-3.3-70B & $6.9 \pm 4.2$ & $17.0 \pm 6.9$ & $0.235 \pm 0.048$ & $0.298 \pm 0.052$ & $0.1 \pm 0.4$ \\
  & & Mixtral-8x22B & $6.1 \pm 3.7$ & $15.0 \pm 6.2$ & $0.310 \pm 0.039$ & $0.416 \pm 0.042$ & $1.9 \pm 1.5$ \\
  \cmidrule{2-8}
  & \multirow{3}{*}{LLM+Premise(Hybrid)} & GPT-4o & $10.2 \pm 5.9$ & $23.1 \pm 8.4$ & $0.163 \pm 0.032$ & $0.228 \pm 0.043$ & $0.0 \pm 0.0$ \\
  & & Llama-3.3-70B & $9.2 \pm 5.1$ & $20.6 \pm 7.8$ & $0.212 \pm 0.045$ & $0.273 \pm 0.052$ & $0.1 \pm 0.4$ \\
  & & Mixtral-8x22B & $6.6 \pm 4.1$ & $15.9 \pm 6.4$ & $0.275 \pm 0.043$ & $0.378 \pm 0.053$ & $2.5 \pm 2.3$ \\

\specialrule{1.2pt}{1pt}{3pt}

\multirow{9}{*}{\rotatebox[origin=c]{90}{\textbf{Prob. Insights}}} 
  & \multirow{3}{*}{\makecell[l]{LLM+Premise(Vector)\\+Insights}} & GPT-4o & $24.7 \pm 11.4$ & $39.9 \pm 12.1$ & $0.128 \pm 0.042$ & $0.198 \pm 0.054$ & $0.0 \pm 0.0$ \\
  & & Llama-3.3-70B & $10.7 \pm 6.3$ & $24.6 \pm 10.6$ & $0.205 \pm 0.047$ & $0.275 \pm 0.052$ & $0.1 \pm 0.4$ \\
  & & Mixtral-8x22B & $10.2 \pm 4.5$ & $21.5 \pm 5.9$ & $0.277 \pm 0.056$ & $0.395 \pm 0.062$ & $1.0 \pm 1.5$ \\
  \cmidrule{2-8}
  & \multirow{3}{*}{\makecell[l]{LLM+Premise(BM25)\\+Insights}}& GPT-4o & $23.3 \pm 10.1$ & $37.6 \pm 11.5$ & $0.132 \pm 0.034$ & $0.203 \pm 0.046$ & $0.0 \pm 0.0$ \\
  & & Llama-3.3-70B & $11.3 \pm 6.7$ & $24.1 \pm 11.5$ & $0.203 \pm 0.044$ & $0.272 \pm 0.047$ & $0.1 \pm 0.4$ \\
  & & Mixtral-8x22B & $10.2 \pm 4.9$ & $20.9 \pm 7.9$ & $0.290 \pm 0.054$ & $0.410 \pm 0.059$ & $0.8 \pm 1.2$ \\
  \cmidrule{2-8}
  & \multirow{3}{*}{\makecell[l]{LLM+Premise(Hybrid)\\+Insights}} & GPT-4o & $23.9 \pm 10.7$ & $39.2 \pm 11.9$ & $0.127 \pm 0.037$ & $0.197 \pm 0.049$ & $0.0 \pm 0.0$ \\
  & & Llama-3.3-70B & $11.6 \pm 6.3$ & $25.0 \pm 10.9$ & $0.201 \pm 0.046$ & $0.267 \pm 0.053$ & $0.1 \pm 0.4$ \\
  & & Mixtral-8x22B & $10.9 \pm 5.3$ & $23.4 \pm 7.2$ & $0.255 \pm 0.046$ & $0.373 \pm 0.049$ & $0.8 \pm 1.4$ \\

\specialrule{1.2pt}{1pt}{3pt}

\multirow{3}{*}{\rotatebox[origin=c]{90}{\textbf{}}} 
  & \multirow{3}{*}{Auto-BN} & GPT-4o & $\mathbf{28.0 \pm 15.5}$ & $\mathbf{46.7 \pm 16.2}$ & $\mathbf{0.107 \pm 0.043}$ & $0.169 \pm 0.058$ & $0.0 \pm 0.0$ \\
  & & Llama-3.3-70B & $25.4 \pm 13.6$ & $44.8 \pm 15.0$ & $0.109 \pm 0.040$ & $\mathbf{0.168 \pm 0.053}$ & $0.0 \pm 0.0$ \\
  & & Mixtral-8x22B & $26.1 \pm 15.4$ & $44.6 \pm 16.1$ & $0.112 \pm 0.046$ & $0.174 \pm 0.062$ & $0.0 \pm 0.2$ \\

\bottomrule
\end{tabular}
}
\caption{Evaluation across factual and probabilistic reasoning strategies across different LLMs.}
\label{table:main-results}
\end{table*}

\subsection{Baselines}

We compare our framework against a diverse set of baselines that reflect factual and probabilistic approaches to probabilistic QA over tabular data:

\noindent \textbf{NL2SQL}: Apply recent NL2SQL work \cite{gao2023text} in probabilistic setting, where the model generates a SQL query from the question, executes it on the database, and uses the result to infer a probabilistic answer.

\noindent \textbf{LLM + Table}: A direct prompting strategy where the question is answered using the table description and a random sample of rows (until reaching the maximum context window size), without structured reasoning or symbolic modeling.

\noindent \textbf{Reasoning Models}: Same strategy as LLM + Table, but using advanced reasoning models and their inherently internal CoT.

\noindent \textbf{TableRAG}: TableRAG \cite{chen2024tablerag} is a large table understanding approach by LLM with retrieval-augmented generation strategy.

\noindent \textbf{LLM + Premise}: This baseline resembles CausalCoT~\citep{jin2023cladder} extended with large-scale probabilistic premise retrieval. It identifies nodes and states from the question using LLM, retrieves probabilistic premises, then uses LLM to answer the question with all information. Three premise retrieval strategies (Appendix~\ref{appendix:premisestore}) are evaluated.

\noindent \textbf{LLM + Premise + Insights}: Extends the previous approach by also incorporating a small set (20) of high-impact insights generated by our proposed Markov-Blanket-based insight generation approach (see Appendix~\ref{appendix:baselines} for details).

\noindent \textbf{Auto-BN:}
Our proposed fully automated neuro-symbolic pipeline (\S\ref{sec-method}). To ensure a strict, real-world evaluation, Auto-BN does not access the human-verified Gold BNs, premises, or insights provided by the benchmark. 

Table~\ref{tab:artifact-summary} summarizes the benchmark artifacts available to each baseline. Complete implementation details for all baselines, including prompt templates, premise storage formats, and retrieval strategies (Vector, BM25, Hybrid), are in Appendix~\ref{appendix:baselines}. Appendix~\ref{appendix:gold-bn} replaces the induced BN with the expert-validated Gold BN (83.8\% vs.\ 46.7\% $\text{Acc}_{0.05}$ with GPT-4o) and shows low directed overlap between the two (F1 of 0.159).

\subsection{Main Results}

Table~\ref{table:main-results} reports the performance of all evaluated methods across various configurations\footnotemark. We report the mean and standard deviation of each metric across all benchmark datasets (tables), computed per $(\text{question type}, \text{baseline})$ pair. The standard deviation reflects performance variability across 10 datasets for the same question type and method. Our results highlight several key observations:

\footnotetext{Each result corresponds to a single run per method per dataset. Reported mean and standard deviation are computed across benchmark datasets for each question type and method.}

\textbf{Factual QA Methods.}
While LLMs prompted with raw table rows (LLM+Table) achieve reasonably low MAE and RMSE, their accuracy in predicting precise probabilities ($\text{Acc}_{x}$) is limited. These models rely on memorization and pattern-matching from examples but lack the structured grounding necessary for calibrated inference. Moreover, this performance is achieved using extra-long contexts, making such approaches computationally expensive and impractical at scale. Even when utilizing advanced reasoning models (GPT-5.4, DeepSeek-R1), direct probabilistic derivation based on raw tables remains unsatisfactory, indicating that internal multi-step reasoning cannot easily substitute for exact statistical aggregation.

Despite being designed for large-table understanding, TableRAG  underperforms even the basic LLM+Table baseline (7.6\% vs.\ 11.1\% $\text{Acc}_{0.02}$ with GPT-4o). This exposes a fundamental mismatch: TableRAG optimizes for localized fact lookup (``needles in a haystack''). By retrieving a bounded set of discrete cell values without their underlying frequency statistics, it cannot aggregate the global evidence required to reconstruct the dense marginal and conditional distributions necessary for probabilistic reasoning.

NL2SQL-style models perform poorly in this setting, as expected. They frequently fail to produce valid SQL under zero-shot prompting, especially Mixtral. More fundamentally, deterministic query execution cannot express conditional uncertainty. For instance, retrieving zero matched records does not equate to a zero probability---highlighting a core limitation of factual QA pipelines when applied to probabilistic tasks.

\textbf{Premise-Based Inference:}
Inspired by approaches like CLADDER that use LLMs for one-hop probabilistic reasoning, we evaluated whether our benchmark could support similar derivations. However, due to the significantly larger and more complex premise sets in \benchmarkname (79k premises per table), LLMs struggled to combine the relevant evidence into coherent probabilistic derivations.
Among premise retrievers, vector-based retrieval consistently outperformed BM25. Hybrid retrieval (combining both methods) yields marginal improvement over individual retrieval.

\textbf{Incorporating Insights}:
Across all retrieval strategies, augmenting retrieved premises with a compact set of 20 high-impact insights substantially improved performance. By distilling the large premise space into a concise and informative subset, insights reduce cognitive overhead, improve answer calibration, and mitigate hallucinations in downstream reasoning.

\textbf{Auto-BN}:
Auto-BN substantially outperforms all baselines across core metrics. With GPT-4o, it achieves 28.0\% $\text{Acc}_{0.02}$ and MAE of 0.107, while successfully maintaining a near-zero error rate. This demonstrates the effectiveness of integrating probabilistic graphical model reasoning with robust LLM-based query understanding. By learning BNs directly from data and executing fail-safe semantic parsing, Auto-BN provides highly accurate answers grounded in exact probabilistic inference.

\textbf{Agentic Extension (Auto-BN-Tool).} 
To somehow relax the limitations of the pipelined approach and  allow LLM-based self recovery, we extend \textbf{Auto-BN} into a callable tool for multi-turn LLM agents. This formulation enables the LLM to autonomously invoke probability calculations, iteratively decompose compositional queries, and synthesize comprehensive answers from multiple probabilistic sub-steps. We ran Auto-BN-Tool with GPT-4o and its performance is very close to Auto-BN (more details in Appendix~\ref{appendix:autobn_tool}).

\textbf{Probability Hallucination vs.\ Grounded Evidence:} 
While LLM baselines artificially achieve near-zero error rates by forcing formatted outputs, they frequently hallucinate uniform priors when uncertain (e.g., predicting 0.2 for 5 candidate states). Auto-BN strictly precludes this by grounding every prediction in exact Bayesian inference. Its own near-zero error rate stems from robust query translation, execution-guided refinement and semantic relaxation rather than blind guessing, guaranteeing a transparent, verifiable reasoning trace.

Further experimental results, including performance breakdowns, running time, multi-hop premises and ablation study are in Appendix~\ref{sec-addiexp}.

\subsection{Effect of Number of Premises}

\begin{figure}[ht]
   \centering
   \includegraphics[width=0.9\columnwidth]{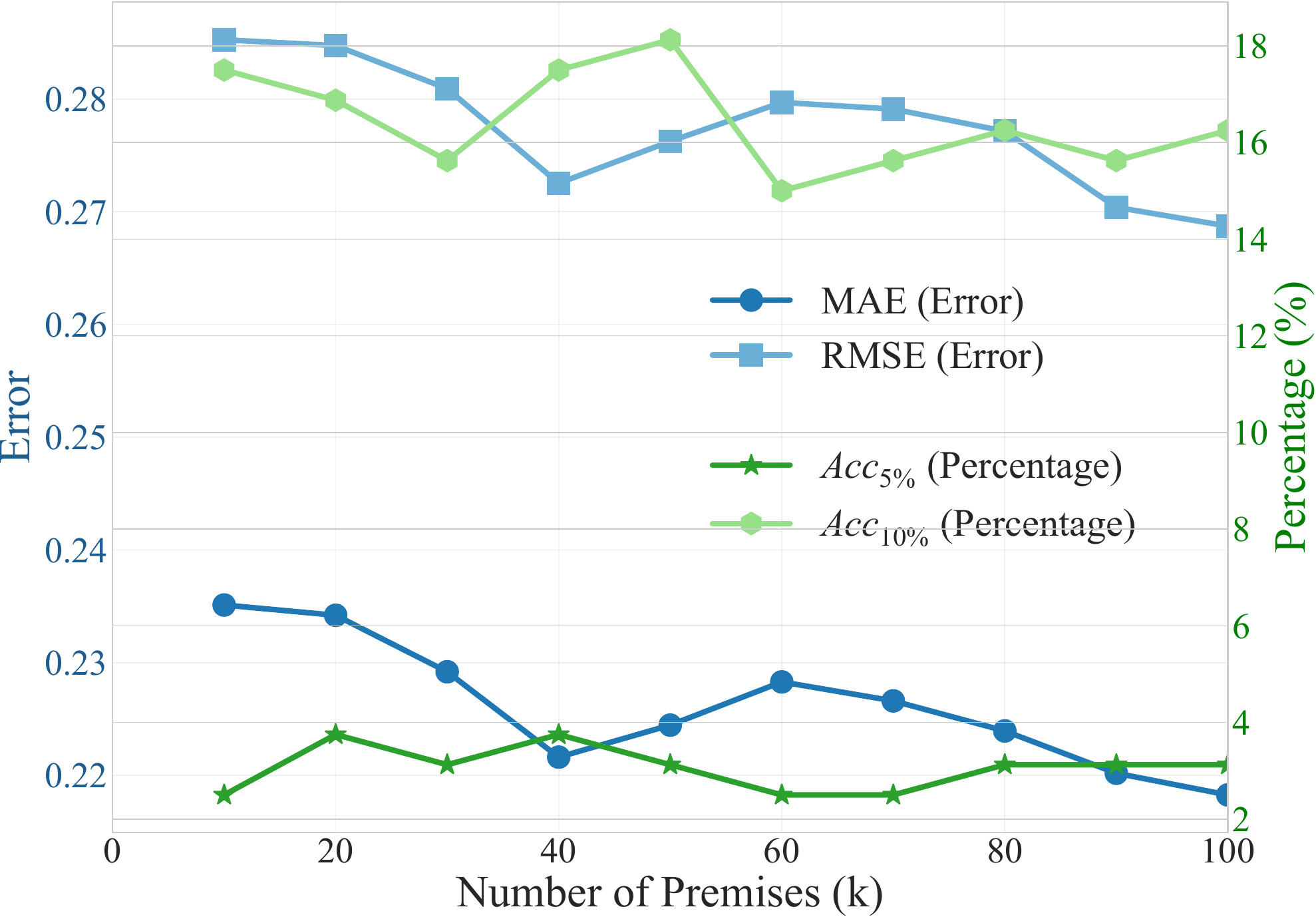}
   \caption{Impact of \# premises on performance}
   \label{fig:num_premises}
\end{figure}

To study the effect of premise count, we conduct experiments on the complex \textit{hockey scoring} dataset using the LLM + Premises setup. As shown in Figure~\ref{fig:num_premises}, increasing the number of retrieved premises yields marginal improvements in MAE and RMSE, but does not consistently boost accuracy. This suggests diminishing returns, where additional premises may introduce redundancy or noise. In contrast, supplementing with a small set of high-impact insights yields in more substantial performance gains, as shown in Table \ref{table:main-results}.

\section{Conclusion}\label{sec-conclusion}

We introduce \benchmarkname, a benchmark for probabilistic question answering over real-world tabular data, and Auto-BN, a framework for answering such questions. Unlike prior work that relies on small, curated Bayesian Networks or synthetic templates, \benchmarkname supports large-scale, data-driven reasoning by inducing and human-validating Bayesian Networks from relational tables. It includes diverse query types, natural language variations, and structured artifacts such as premises and insights. Auto-BN combines probabilistic inference with robust LLM-based query translation. Experimental results show that our framework outperforms retrieval-based and probability-based baselines, highlighting the value of integrating probabilistic graphical models with neural models for reasoning under uncertainty.

\section*{Limitations}\label{sec:limitations}
While our framework effectively learns Bayesian network structures from tabular records, it operates in a purely data-driven manner and does not incorporate external knowledge or semantic priors beyond human validation of the benchmark artifacts. Consequently, the learned statistical dependencies are limited to patterns present in the observed data. 

We emphasize that our framework evaluates \textit{observational probabilistic reasoning}. While we occasionally employ terminology from the causal discovery literature (e.g., the PC algorithm) to compare with related work, we do not claim the discovery of true causal mechanisms or the estimation of interventional effects (do-calculus). Some directed edges in the induced Bayesian Networks represent strict statistical correlations (Markov Equivalence) rather than verified causal mechanisms. 

Finally, while we introduced a new benchmark and demonstrated strong baseline methods, we do not claim our proposed Auto-BN pipeline to be the optimal solution for this benchmark. The benchmark is intended as a platform to foster further exploration at the intersection of tabular reasoning, uncertainty quantification, and probabilistic modeling.

\section*{Ethical Considerations}
\label{sec-ethic}
Our framework automatically fit given tabular data across domains, which may propagate underlying biases in the data. The learned structures reflect statistical estimations based solely on observed records, without incorporating broader world knowledge. As such, they may not represent ground truth and should not be used as generic justification or reliable decision making, such as investigations or healthcare applications.

\section*{Acknowledgments}
The authors thank Sajjadur Rahman for his contributions to an earlier version of this work. As he was not involved in the substantial revisions and preparation of the present manuscript, he chose not to be included as an author of this version.

\bibliography{custom,anthology-1,anthology-2}

\begin{thebibliography}{32}
\providecommand{\natexlab}[1]{#1}

\bibitem[{Ankan and Textor(2024)}]{Ankan2024}
Ankur Ankan and Johannes Textor. 2024.
\newblock \href {http://jmlr.org/papers/v25/23-0487.html} {pgmpy: A python
  toolkit for bayesian networks}.
\newblock \emph{Journal of Machine Learning Research}, 25(265):1--8.

\bibitem[{Bondarenko et~al.(2022)Bondarenko, Wolska, Heindorf, Bl{\"u}baum,
  Ngomo, Stein, Braslavski, Hagen, and Potthast}]{bondarenko2022causalqa}
Alexander Bondarenko, Magdalena Wolska, Stefan Heindorf, Lukas Bl{\"u}baum,
  Axel-Cyrille~Ngonga Ngomo, Benno Stein, Pavel Braslavski, Matthias Hagen, and
  Martin Potthast. 2022.
\newblock Causalqa: A benchmark for causal question answering.
\newblock In \emph{Proceedings of the 29th International Conference on
  Computational Linguistics}, pages 3296--3308.

\bibitem[{Chakraborty et~al.(2025)Chakraborty, Ornik, and
  Driggs-Campbell}]{chakraborty2025hallucination}
Neeloy Chakraborty, Melkior Ornik, and Katherine Driggs-Campbell. 2025.
\newblock Hallucination detection in foundation models for decision-making: A
  flexible definition and review of the state of the art.
\newblock \emph{ACM Computing Surveys}.

\bibitem[{Chen et~al.(2024)Chen, Miculicich, Eisenschlos, Wang, Wang, Chen,
  Fujii, Lin, Lee, and Pfister}]{chen2024tablerag}
Si-An Chen, Lesly Miculicich, Julian~Martin Eisenschlos, Zifeng Wang, Zilong
  Wang, Yanfei Chen, Yasuhisa Fujii, Hsuan-Tien Lin, Chen-Yu Lee, and Tomas
  Pfister. 2024.
\newblock Tablerag: Million-token table understanding with language models.
\newblock In \emph{Advances in Neural Information Processing Systems},
  volume~37.

\bibitem[{De~Raedt et~al.(2007)De~Raedt, Kimmig, and Toivonen}]{de2007problog}
Luc De~Raedt, Angelika Kimmig, and Hannu Toivonen. 2007.
\newblock Problog: A probabilistic prolog and its application in link
  discovery.
\newblock In \emph{IJCAI 2007, Proceedings of the 20th international joint
  conference on artificial intelligence}, pages 2462--2467. IJCAI-INT JOINT
  CONF ARTIF INTELL.

\bibitem[{Deng et~al.(2022)Deng, Sun, Lees, Wu, and Yu}]{deng2022turl}
Xiang Deng, Huan Sun, Alyssa Lees, You Wu, and Cong Yu. 2022.
\newblock Turl: Table understanding through representation learning.
\newblock \emph{ACM SIGMOD Record}, 51(1):33--40.

\bibitem[{Gao et~al.(2023)Gao, Wang, Li, Sun, Qian, Ding, and
  Zhou}]{gao2023text}
Dawei Gao, Haibin Wang, Yaliang Li, Xiuyu Sun, Yichen Qian, Bolin Ding, and
  Jingren Zhou. 2023.
\newblock Text-to-sql empowered by large language models: A benchmark
  evaluation.
\newblock \emph{arXiv preprint arXiv:2308.15363}.

\bibitem[{Grattafiori et~al.(2024)Grattafiori, Dubey, Jauhri, Pandey, Kadian
  et~al.}]{grattafiori2024llama3}
Aaron Grattafiori, Abhimanyu Dubey, Abhinav Jauhri, Abhinav Pandey, Abhishek
  Kadian, and 1 others. 2024.
\newblock The {Llama} 3 herd of models.
\newblock \emph{arXiv preprint arXiv:2407.21783}.

\bibitem[{Herzig et~al.(2020)Herzig, Nowak, Mueller, Piccinno, and
  Eisenschlos}]{herzig2020tapas}
Jonathan Herzig, Pawel~Krzysztof Nowak, Thomas Mueller, Francesco Piccinno, and
  Julian Eisenschlos. 2020.
\newblock Tapas: Weakly supervised table parsing via pre-training.
\newblock In \emph{Proceedings of the 58th Annual Meeting of the Association
  for Computational Linguistics}, pages 4320--4333.

\bibitem[{Huang et~al.(2025)Huang, Yu, Ma, Zhong, Feng, Wang, Chen, Peng, Feng,
  Qin et~al.}]{huang2025survey}
Lei Huang, Weijiang Yu, Weitao Ma, Weihong Zhong, Zhangyin Feng, Haotian Wang,
  Qianglong Chen, Weihua Peng, Xiaocheng Feng, Bing Qin, and 1 others. 2025.
\newblock A survey on hallucination in large language models: Principles,
  taxonomy, challenges, and open questions.
\newblock \emph{ACM Transactions on Information Systems}, 43(2):1--55.

\bibitem[{Hurst et~al.(2024)Hurst, Lerer, Goucher, Perelman, Ramesh, Clark,
  Ostrow, Welihinda, Hayes, Radford et~al.}]{hurst2024gpt}
Aaron Hurst, Adam Lerer, Adam~P Goucher, Adam Perelman, Aditya Ramesh, Aidan
  Clark, AJ~Ostrow, Akila Welihinda, Alan Hayes, Alec Radford, and 1 others.
  2024.
\newblock Gpt-4o system card.
\newblock \emph{arXiv preprint arXiv:2410.21276}.

\bibitem[{Jiang et~al.(2024)Jiang, Sablayrolles, Roux, Mensch, Savary, Bamford,
  Chaplot, Casas, Hanna, Bressand et~al.}]{jiang2024mixtral}
Albert~Q Jiang, Alexandre Sablayrolles, Antoine Roux, Arthur Mensch, Blanche
  Savary, Chris Bamford, Devendra~Singh Chaplot, Diego de~las Casas, Emma~Bou
  Hanna, Florian Bressand, and 1 others. 2024.
\newblock Mixtral of experts.
\newblock \emph{arXiv preprint arXiv:2401.04088}.

\bibitem[{Jin et~al.(2023)Jin, Chen, Leeb, Gresele, Kamal, Lyu, Blin,
  Gonzalez~Adauto, Kleiman-Weiner, Sachan et~al.}]{jin2023cladder}
Zhijing Jin, Yuen Chen, Felix Leeb, Luigi Gresele, Ojasv Kamal, Zhiheng Lyu,
  Kevin Blin, Fernando Gonzalez~Adauto, Max Kleiman-Weiner, Mrinmaya Sachan,
  and 1 others. 2023.
\newblock Cladder: Assessing causal reasoning in language models.
\newblock \emph{Advances in Neural Information Processing Systems},
  36:31038--31065.

\bibitem[{Kalisch and B{\"u}hlman(2007)}]{kalisch2007estimating}
Markus Kalisch and Peter B{\"u}hlman. 2007.
\newblock Estimating high-dimensional directed acyclic graphs with the
  pc-algorithm.
\newblock \emph{Journal of Machine Learning Research}, 8(3).

\bibitem[{Li et~al.(2024)Li, Luo, Chai, Li, and Tang}]{li2024dawn}
Boyan Li, Yuyu Luo, Chengliang Chai, Guoliang Li, and Nan Tang. 2024.
\newblock The dawn of natural language to sql: are we fully ready?
\newblock \emph{arXiv preprint arXiv:2406.01265}.

\bibitem[{Li et~al.(2026)Li, Liu, Song, Chi, Shi, Zhao, Chang, Wang, Yang,
  Wang, Deng, and Feng}]{li2026treb}
Ce~Li, Xiaofan Liu, Zhiyan Song, Ce~Chi, Boshen Shi, Chen Zhao, Guanguang
  Chang, Zhendong Wang, Kexin Yang, Xing Wang, Chao Deng, and Junlan Feng.
  2026.
\newblock \href {https://doi.org/10.1145/3805712.3808617} {{TReB}: A
  comprehensive benchmark for evaluating table reasoning capabilities of large
  language models}.
\newblock In \emph{Proceedings of the 49th International ACM SIGIR Conference
  on Research and Development in Information Retrieval}, pages 3267--3275.
  Association for Computing Machinery.

\bibitem[{Li et~al.(2023)Li, Hui, Qu, Yang, Li, Li, Wang, Qin, Geng, Huo
  et~al.}]{li2024can}
Jinyang Li, Binyuan Hui, Ge~Qu, Jiaxi Yang, Binhua Li, Bowen Li, Bailin Wang,
  Bowen Qin, Ruiying Geng, Nan Huo, and 1 others. 2023.
\newblock \href {https://doi.org/10.52202/075280-1835} {Can {LLM} already serve
  as a database interface? {A} {BI}g bench for large-scale database grounded
  text-to-{SQLs}}.
\newblock In \emph{Advances in Neural Information Processing Systems},
  volume~36, pages 42330--42357.

\bibitem[{Liu(2022)}]{Liu_LlamaIndex_2022}
Jerry Liu. 2022.
\newblock {LlamaIndex}.
\newblock \url{https://github.com/run-llama/llama_index}.
\newblock Software.

\bibitem[{Liu et~al.(2024)Liu, Wu, Wu, Lu, Chang, and Feng}]{liu2024llms}
Xiao Liu, Zirui Wu, Xueqing Wu, Pan Lu, Kai-Wei Chang, and Yansong Feng. 2024.
\newblock Are llms capable of data-based statistical and causal reasoning?
  benchmarking advanced quantitative reasoning with data.
\newblock In \emph{Findings of the Association for Computational Linguistics
  ACL 2024}, pages 9215--9235.

\bibitem[{{Meta}(2024)}]{meta2024llama33}
{Meta}. 2024.
\newblock {Llama 3.3} model card.
\newblock
  \url{https://github.com/meta-llama/llama-models/blob/main/models/llama3_3/MODEL_CARD.md}.

\bibitem[{{Mistral AI}(2024)}]{mistral2024mixtral8x22b}
{Mistral AI}. 2024.
\newblock Cheaper, better, faster, stronger.
\newblock \url{https://mistral.ai/news/mixtral-8x22b}.
\newblock Blog post announcing {Mixtral 8x22B}, 17 April 2024.

\bibitem[{Nafar et~al.(2025)Nafar, Venable, and
  Kordjamshidi}]{nafar2025reasoning}
Aliakbar Nafar, Kristen~Brent Venable, and Parisa Kordjamshidi. 2025.
\newblock Reasoning over uncertain text by generative large language models.
\newblock In \emph{Proceedings of the AAAI Conference on Artificial
  Intelligence}, volume~39, pages 24911--24920.

\bibitem[{Pearl(1988)}]{pearl1988}
Judea Pearl. 1988.
\newblock \emph{Probabilistic Reasoning in Intelligent Systems: Networks of
  Plausible Inference}.
\newblock Morgan Kaufmann Publishers Inc., San Francisco, CA, USA.

\bibitem[{Pourreza et~al.(2024)Pourreza, Li, Sun, Chung, Talaei, Kakkar, Gan,
  Saberi, Ozcan, and Arik}]{pourreza2024chase}
Mohammadreza Pourreza, Hailong Li, Ruoxi Sun, Yeounoh Chung, Shayan Talaei,
  Gaurav~Tarlok Kakkar, Yu~Gan, Amin Saberi, Fatma Ozcan, and Sercan~O Arik.
  2024.
\newblock Chase-sql: Multi-path reasoning and preference optimized candidate
  selection in text-to-sql.
\newblock \emph{arXiv preprint arXiv:2410.01943}.

\bibitem[{Reimers and Gurevych(2019)}]{reimers-2019-sentence-bert}
Nils Reimers and Iryna Gurevych. 2019.
\newblock \href {https://arxiv.org/abs/1908.10084} {Sentence-bert: Sentence
  embeddings using siamese bert-networks}.
\newblock In \emph{Proceedings of the 2019 Conference on Empirical Methods in
  Natural Language Processing}. Association for Computational Linguistics.

\bibitem[{Schrader et~al.(2024)Schrader, Lange, Razniewski, and
  Friedrich}]{schrader-etal-2024-quite}
Timo~Pierre Schrader, Lukas Lange, Simon Razniewski, and Annemarie Friedrich.
  2024.
\newblock \href {https://doi.org/10.18653/v1/2024.emnlp-main.153} {{QUITE}:
  Quantifying uncertainty in natural language text in {B}ayesian reasoning
  scenarios}.
\newblock In \emph{Proceedings of the 2024 Conference on Empirical Methods in
  Natural Language Processing}, pages 2634--2652, Miami, Florida, USA.
  Association for Computational Linguistics.

\bibitem[{Shen et~al.(2024)Shen, Wang, Rahman, and
  Kandogan}]{10.1145/3627673.3679216}
Chen Shen, Jin Wang, Sajjadur Rahman, and Eser Kandogan. 2024.
\newblock \href {https://doi.org/10.1145/3627673.3679216} {Demonstration of a
  multi-agent framework for text to sql applications with large language
  models}.
\newblock In \emph{Proceedings of the 33rd ACM International Conference on
  Information and Knowledge Management}, CIKM '24, page 5280–5283, New York,
  NY, USA. Association for Computing Machinery.

\bibitem[{Shrestha et~al.(2025)Shrestha, Kim, and
  Ross}]{shrestha2025mathematical}
Safal Shrestha, Minwu Kim, and Keith Ross. 2025.
\newblock Mathematical reasoning in large language models: Assessing logical
  and arithmetic errors across wide numerical ranges.
\newblock \emph{arXiv preprint arXiv:2502.08680}.

\bibitem[{Srivastava et~al.(2023)Srivastava, Rastogi, Rao
  et~al.}]{srivastava2023beyond}
Aarohi Srivastava, Abhinav Rastogi, Abhishek Rao, and 1 others. 2023.
\newblock \href {https://openreview.net/forum?id=uyTL5Bvosj} {Beyond the
  imitation game: Quantifying and extrapolating the capabilities of language
  models}.
\newblock \emph{Transactions on Machine Learning Research}.
\newblock Featured Certification.

\bibitem[{Talaei et~al.(2024)Talaei, Pourreza, Chang, Mirhoseini, and
  Saberi}]{talaei2024chess}
Shayan Talaei, Mohammadreza Pourreza, Yu-Chen Chang, Azalia Mirhoseini, and
  Amin Saberi. 2024.
\newblock Chess: Contextual harnessing for efficient sql synthesis.
\newblock \emph{arXiv preprint arXiv:2405.16755}.

\bibitem[{Wu et~al.(2025)Wu, Yang, Chai, Zhang, Liu, Du, Liang, Shu, Cheng,
  Sun, Li, Li, and Niu}]{wu2025tablebench}
Xianjie Wu, Jian Yang, Linzheng Chai, Ge~Zhang, Jiaheng Liu, Xeron Du,
  Di~Liang, Daixin Shu, Xianfu Cheng, Tianzhen Sun, Tongliang Li, Zhoujun Li,
  and Guanglin Niu. 2025.
\newblock \href {https://doi.org/10.1609/aaai.v39i24.34739} {{TableBench}: A
  comprehensive and complex benchmark for table question answering}.
\newblock In \emph{Proceedings of the AAAI Conference on Artificial
  Intelligence}, volume~39, pages 25497--25506.

\bibitem[{Zhao et~al.(2025)Zhao, Zhang, Wang, Xin, Lu, Li, Lyu, Ou, and
  Song}]{zhao2025tqagent}
Jianbin Zhao, Pengfei Zhang, Yuzhen Wang, Rui Xin, Xiuyuan Lu, Ripeng Li, Shuai
  Lyu, Zhonghong Ou, and Meina Song. 2025.
\newblock Tqagent: Enhancing table-based question answering with knowledge
  graphs and tree-structured reasoning.
\newblock \emph{Applied Sciences (2076-3417)}, 15(7).

\end{thebibliography}
\appendix

\section{Benchmark Details}
\label{sec-statistics}
We provide detailed statistics for the \benchmarkname{} benchmark, covering data source selection, the structural properties of the Bayesian Networks, and the distribution of questions across reasoning patterns.

\subsection{Data Source Selection}
\label{subsec:data-source}

\benchmarkname{} is built on 10 real-world relational tables drawn from the BIRD benchmark~\citep{li2024can}.
These 10 tables were carefully selected from hundreds of candidate sources in BIRD according to three criteria that are necessary for constructing a meaningful probabilistic QA benchmark:

\paragraph{Meaningful Probabilistic Dependencies:} The schema must exhibit logical, directional associations between attributes (e.g., \textit{departure delay} influencing \textit{arrival delay}), enabling non-trivial conditional reasoning.

\paragraph{Feature Richness:} The table must feature a high density of categorical or discretizable numerical columns. Databases consisting primarily of unique identifiers, proper names, or unstructured locations were excluded.

\paragraph{Statistical Sufficiency:} The table must contain a sufficient volume of records to ensure that empirical probability distributions and the induced statistical models (e.g., Bayesian Networks) are robust, reliable, and representative of real-world complexity.

The resulting 10 tables span diverse real-world domains, from transportation and finance to sports and social media.
Table~\ref{tab:tables} summarizes the key properties of each source table.

\begin{table*}[t]
\centering
\small
\begin{tabular}{lccc>{\raggedright\arraybackslash}p{5.6cm}}
\toprule
\textbf{Table} & \textbf{Rows} & \textbf{Columns} & \textbf{Domain} & \textbf{Notes} \\
\midrule
airlines & 701,352 & 28 & Transportation & Real-world flight data including delays and routes \\
basketball players & 23,751 & 43 & Sports & Season-level performance statistics for NBA \& ABA players \\
bike stations weather & 3,665 & 24 & Urban Mobility & Bike share stations \& weather data\\
coinmarketcap & 4,441,972 & 19 & Finance & Real-world crypto market history \\
college completion & 3,798 & 62 & Education & US college graduation statistics from IPEDS \\
cookbook nutrition & 878 & 16 & Health & Nutritional info from online recipe collections \\
hockey scoring & 45,967 & 31 & Sports & Professional hockey players statistics per season\\
movies & 4,627 & 13 & Entertainment & Metadata from popular movie databases\\
sales in weather & 20,517 & 20 & Retail & Sales transactions combined with weather info \\
twitter & 99,901 & 14 & Social Media & Tweet metadata sampled over multiple months \\
\midrule
\textbf{Mean} & \textbf{486,061} & \textbf{25.91} & -- & -- \\
\bottomrule
\end{tabular}
\caption{Summary of tables used in experiments}
\label{tab:tables}
\end{table*}

\subsection{Detailed Statistics}
\paragraph{Bayesian Network Properties.} 
As shown in Table~\ref{tab:graph_stats}, the Bayesian Networks induced in \benchmarkname are substantially larger and more structured than those used in prior benchmarks. On average, each network contains over 12 nodes, with approximately 6,000 probabilistic premises per node. This scale highlights a key challenge: methods that require injecting all premises into LLM prompts, such as CLADDER-style symbolic chains, are fundamentally infeasible due to context length limits and reasoning bottlenecks. This motivates our design of insight-driven filtering and structured inference via Auto-BN (\S~\ref{sec-method}).

\begin{table*}[!ht]
\centering
\small
\renewcommand{\arraystretch}{0.95}
\setlength{\tabcolsep}{3pt}
\begin{tabular}{lrrrrrrrrrr}
\toprule
\textbf{Dataset} & \textbf{Nodes} & \textbf{States/Node} & \textbf{Premises} & \textbf{Insights} & \textbf{Prem/Node} & \textbf{Ins/Node} & \textbf{Edges} & \textbf{In-Deg} & \textbf{Out-Deg} & \textbf{Deg} \\
\midrule
airlines & 7 & 5.00 & 5,107 & 130 & 729.6 & 18.6 & 10 & 1.4 & 1.4 & 2.9 \\
basketball players & 19 & 5.00 & 15,568 & 5,652 & 819.4 & 297.5 & 42 & 2.2 & 2.2 & 4.4 \\
bike stations weather & 17 & 4.94 & 23,217 & 1,168 & 1,365.7 & 68.7 & 43 & 2.5 & 2.5 & 5.1 \\
coinmarketcap & 13 & 5.00 & 43,932 & 960 & 3,379.4 & 73.8 & 37 & 2.9 & 2.9 & 5.7 \\
college completion & 15 & 4.67 & 46,337 & 733 & 3,089.1 & 48.9 & 48 & 3.2 & 3.2 & 6.4 \\
cookbook nutrition & 14 & 5.00 & 4,798 & 4,153 & 342.7 & 296.6 & 28 & 2.0 & 2.0 & 4.0 \\
hockey scoring & 13 & 4.92 & 337,559 & 3,063 & 25,966.1 & 235.6 & 41 & 3.2 & 3.2 & 6.3 \\
movies & 6 & 5.00 & 669 & 1,634 & 111.5 & 272.3 & 10 & 1.7 & 1.7 & 3.3 \\
sales in weather & 13 & 5.00 & 305,100 & 4,405 & 23,469.2 & 338.9 & 48 & 3.7 & 3.7 & 7.4 \\
twitter & 8 & 4.88 & 9,331 & 2,220 & 1,166.4 & 277.5 & 19 & 2.4 & 2.4 & 4.8 \\
\midrule
\textbf{Mean} & 12.5 & 4.95 & 79,161.8 & 2,411.8 & 6,043.9 & 192.8 & 32.6 & 2.5 & 2.5 & 5.0 \\
\bottomrule
\end{tabular}
\caption{Statistics of Bayesian Networks for each dataset in \benchmarkname. Degree metrics are averaged per node.}
\label{tab:graph_stats}
\end{table*}

\paragraph{Question Distribution.}
Table~\ref{tab:question_reasoning_distribution} summarizes the distribution of question types across five probabilistic query types. For each probabilistic query, four variations in question formulation are created: \textit{Exact}, \textit{Shifted}, \textit{Certain}, and \textit{Natural}, yielding an equal number of questions per type (table rows). The probabilistic reasoning query types include \textit{forward conditional query} and \textit{diagnostic query}, as well as more complex reasoning patterns such as \textit{interdependent evidence query} and \textit{mixed} queries, with or without redundant context. The benchmark is also balanced across query types (table columns), ensuring broad coverage and structural diversity.

\paragraph{Evaluation Splits.}
Following the protocol of zero-shot benchmarks such as BIG-Bench \cite{srivastava2023beyond}, we treat each table as an independent evaluation unit, with no train/test split. This setup assesses inherent reasoning capabilities without fine-tuning. An optional train/test split is provided for users interested in training/fine-tuning based experiments.

\paragraph{Privacy and Licensing.}
The \benchmarkname{} benchmark does not include names or any information that uniquely identifies individuals. In accordance with its data sources, the benchmark is released under the CC BY-SA 4.0 license.

\begin{table*}[!ht]
\centering
\setlength{\tabcolsep}{3.5pt}
\begin{tabular}{lcccccc}
\toprule
\multirow{3}{*}{\textbf{Question Type}} 
& \multicolumn{5}{c}{\textbf{Probabilistic Query Types}} 
& \multirow{3}{*}{\textbf{Total}} \\
\cmidrule(lr){2-6}
& \shortstack{\textbf{Forward}\\\textbf{Conditional}} 
& \shortstack{\textbf{Diagnostic}} 
& \shortstack{\textbf{Interdependent}\\\textbf{Evidence}} 
& \shortstack{\textbf{Mixed}} 
& \shortstack{\textbf{Mixed with}\\\textbf{Redundant Context}} 
& \\
\midrule
Exact   & 326 & 320 & 304 & 315 & 305 & 1570 \\
Shifted & 326 & 320 & 304 & 315 & 305 & 1570 \\
Certain & 326 & 320 & 304 & 315 & 305 & 1570 \\
Natural & 326 & 320 & 304 & 315 & 305 & 1570 \\
\midrule
\textbf{Total} & 1304 & 1280 & 1216 & 1260 & 1220 & 6280 \\
\bottomrule
\end{tabular}
\caption{Distribution of question types across probabilistic query types (reasoning patterns).}
\label{tab:question_reasoning_distribution}
\end{table*}

\subsection{Examples by Probabilistic Query Type}
Table~\ref{tab:examples-movies} shows one \textit{Natural}-variant question for each probabilistic query type from the \textit{movies} table.

\begin{table*}[t]
\centering
\small
\begin{tabular}{l>{\raggedright\arraybackslash}p{11.1cm}}
\toprule
\textbf{Query Type} & \textbf{Question} \\
\midrule
Forward Conditional & \textit{If we observe that a movie has a short duration and receives below average ratings, how likely is it that the film will generate little to no revenue?} \\
Diagnostic & \textit{Given that the movie has received only a small amount of votes, how likely is it that its popularity is very low?} \\
Interdependent Evidence & \textit{Given that the movie has an extremely high budget, how likely is it to achieve moderate popularity?} \\
Mixed & \textit{Given that a movie has an extended duration, receives a moderate amount of votes, and is produced with a very limited budget, how likely is it that this movie achieves a high average rating?} \\
Mixed with Redundant Context & \textit{Given that a movie has a moderately short duration, receives high ratings, but generates little to no revenue, how likely is it that the movie will have very low popularity?} \\
\bottomrule
\end{tabular}
\caption{One \textit{Natural}-variant question for each probabilistic query type (\textit{movies}).}
\label{tab:examples-movies}
\end{table*}

\subsection{Worked Example}\label{app:worked-example}
This example traces one \benchmarkname{} instance from the input table to the gold answer.

\textbf{Input table.} The \textit{movies} table has 4,627 rows and 13 columns (Table~\ref{tab:tables}), including runtime, vote\_average, vote\_count, and popularity; a typical row has runtime 105, vote\_average 7.3, vote\_count 959, and popularity 23.7.

\textbf{BN.} Figure~\ref{fig:causal_graph_example} shows the Gold BN, whose nodes are the discretized columns.

\textbf{Question.} \textit{``Given that a movie has a runtime of 103 minutes and an average vote of 8.9, what is the probability that its vote count is between 750 and 3000?''} (\textit{Certain} variant).

\textbf{BN query.} $P(\textit{vote\_count}{\in}(750,3000] \mid \textit{runtime}{\in}(100,110], \textit{vote\_average}{\in}(7.0,10.0])$: each stated value is mapped to the bin that contains it.

\textbf{Inference path.} Exact inference propagates the evidence over two hops, vote\_average $\rightarrow$ popularity $\rightarrow$ vote\_count.

\textbf{Answer.} The probability rises from the prior $P(\textit{vote\_count}{\in}(750,3000]) = 0.209$ to the gold answer of 0.352.

\section{Premise Store and Retrieval} 
\label{appendix:premisestore}

To enable retrieval on large scale premises, for each premise, \benchmarkname provides two types of premises stores: the Numeric Premise Store stores the nodes and states in numeric format (e.g. "\textit{price increases from 1\% to 5\%}"), and the Natural Premise Store stores in natural language text format ("\textit{price slightly increases}").

We designed 3 types of premise store retrievers base on LlamaIndex \cite{Liu_LlamaIndex_2022}:

\begin{compactitem}
\item \textbf{Vector} Retrieve premises according to text encoding SentenceTransformer \cite{reimers-2019-sentence-bert} similarity.
\item \textbf{BM25} Retrieve premises using the BM25 (Best Match 25) method that effectively ranks premises based on query term occurrence and rarity across the corpus.
\item \textbf{Hybrid} Retrieve from both Vector and BM25 Retriever, combine and rerank the results.
\end{compactitem}

\section{Baselines Implementation Details}
\label{appendix:baselines}

We compare our Auto-BN framework against a diverse set of baselines spanning LLM-only, retrieval-augmented, and symbolic approaches. Below we describe the artifacts summary and implementation details referenced in \S~\ref{sec-exp}.

\subsection{Artifacts Summary}
\begin{figure}[!ht]
   \centering
   \includegraphics[width=0.5\textwidth]{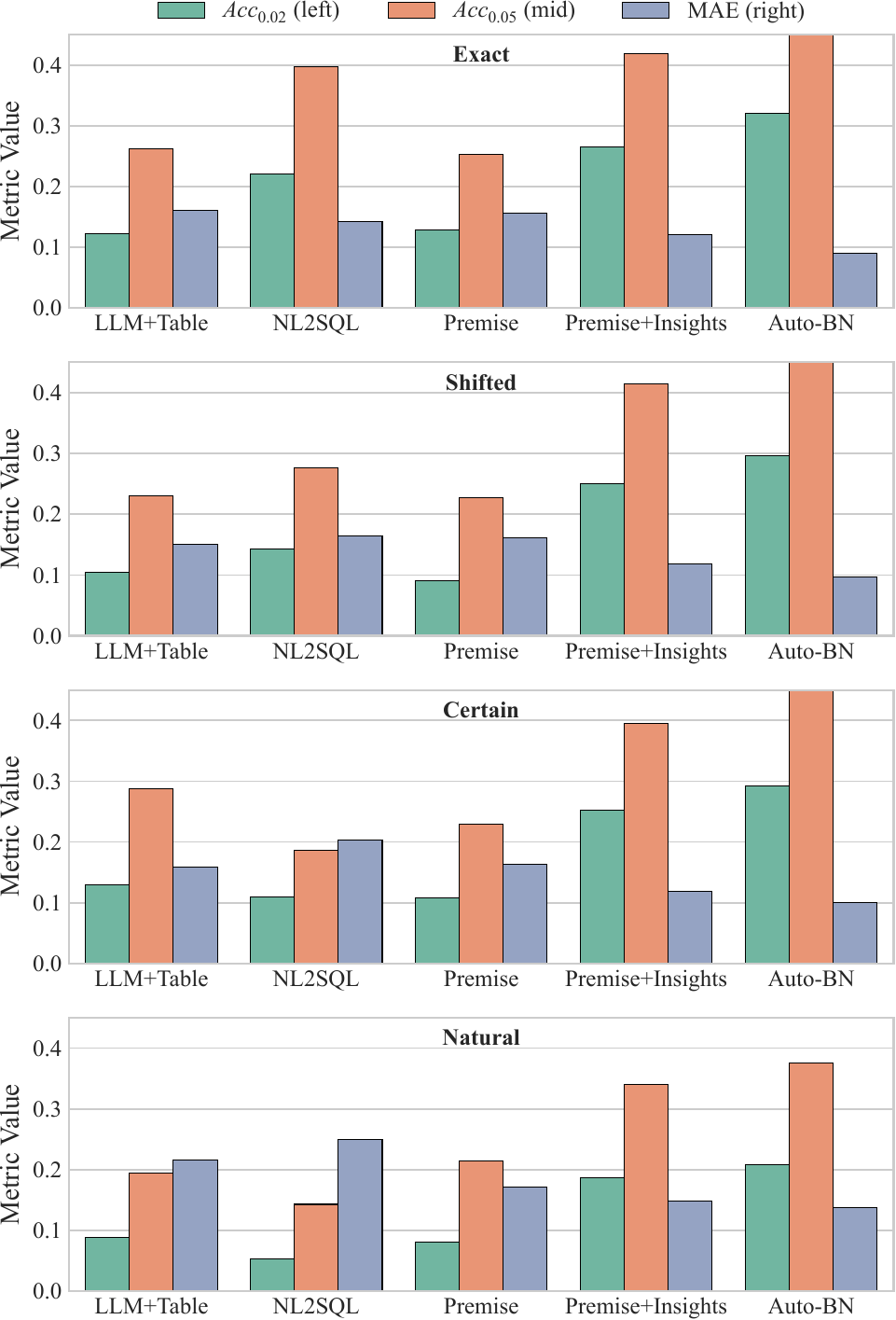}
   \caption{Performance of baselines by question type. Bars represent $Acc_{0.02}$ (↑), $Acc_{0.05}$ (↑), and MAE (↓) for each method.}
   \label{fig:metrics_question_type}
\end{figure}

Table~\ref{tab:artifact-summary} summarizes the benchmark artifacts used by each method, making the information access conditions explicit for direct comparison. No baselines utilize the ground-truth BN.

\begin{table*}[!ht]
\centering
\setlength{\tabcolsep}{4pt}
\begin{tabular}{lcccc}
\toprule
\textbf{Method} & \textbf{Raw Table} & \textbf{Ground-Truth BN} & \textbf{Premises} & \textbf{Insights} \\
\midrule
LLM + Table         & \checkmark &            &            &            \\
NL2SQL              & \checkmark &            &            &            \\
TableRAG            & \checkmark &            &            &            \\
Reasoning Model     & \checkmark &            &            &            \\
LLM+Premise         &            &            & \checkmark &            \\
LLM+Premise+Insights &           &            & \checkmark & \checkmark \\
Auto-BN             & \checkmark &            &            &            \\
\midrule
\textit{Auto-BN (Gold)*} & \textit{\checkmark} & \textit{\checkmark} & & \\
\bottomrule
\end{tabular}
\caption{Benchmark artifacts used by each method. All baselines utilize the table description to understand the natural language question. Premises and Insights are atomic minimum one-hop reasoning instances derived from the human-verified Ground-Truth BN provided by the benchmark. Auto-BN relies exclusively on the Raw Table to induce its own probabilistic model. (*\textit{Auto-BN (Gold)} is an upper-bound ablation evaluated in Appendix E to isolate query translation performance).}
\label{tab:artifact-summary}
\end{table*}

\subsection{Auto-BN Evaluation}\label{appendix:autobn}
While the technical mechanisms of the Auto-BN pipeline are detailed in \S\ref{sec-method}, we emphasize a critical aspect of its evaluation protocol here. In our main experiments, Auto-BN operates in a strictly automated, \textit{structure-free} setting. It accesses only the raw tables and their descriptions, without any access to the ground-truth BNs. Consequently, the data-specific Bayesian Networks induced by Auto-BN act purely as observational probabilistic representations of the underlying tabular data, and do not benefit from the domain experts' manual refinement or injection of semantic knowledge. 

By contrast, the premise- and insight-augmented baselines leverage artifacts derived directly from the expert-curated ground-truth BNs. We intentionally adopt this asymmetric evaluation protocol to prevent structural data leakage and rigorously test Auto-BN's fully automated induction capabilities. For an analysis of Auto-BN's upper-bound performance when provided with the human-verified ground-truth BNs---which isolates query translation effectiveness from structure induction challenges---please refer to the ablation study in Appendix~\ref{appendix:gold-bn}.

\subsection{NL2SQL (Factual Retrieval)}
This baseline tests whether deterministic SQL factual retrieval can support probabilistic inference. Following \citet{gao2023text}, we apply a text-to-SQL model that generates a SQL query from the natural language question, using the database schema (same with \citet{gao2023text}) and table description. The result of executing this query (e.g., a count, average, or conditionally filtered subset) is then passed back to the LLM alongside the original question to estimate the desired probability. This tests whether factual retrieval systems can be repurposed for uncertainty estimation, despite lacking probabilistic semantics.

\subsection{LLM + Table (Neural-Only)}
This baseline probes the LLM’s ability to estimate probabilities directly from examples without structured reasoning. Inspired by the BIRD benchmark~\citep{li2024can}, we concatenate the table schema, table description and table rows with the question, and prompt the LLM for a probability estimate. In the actual situation, due to the large token size of each record, we iteratively append rows to prompt until reach LLM context window limit or 16,384 tokens. This approach avoids any symbolic structure or inference, relying purely on learned statistical patterns in the data. 

\subsection{TableRAG (Retrieval-Augmented Neural)}\label{subsec:tablerag_details}

To evaluate state-of-the-art retrieval-augmented generation on tabular data, we implement the TableRAG baseline \citep{chen2024tablerag} using the official Google Research codebase. Unlike the \textit{LLM + Table} baseline, which relies on row sampling to fill the context window, TableRAG leverages query expansion and dense cell retrieval to dynamically extract the most semantically relevant tabular data.

To accommodate the massive scale of the \benchmarkname{} datasets (which average 486k rows and scale up to 4.4M rows), we made two necessary adaptations to the default TableRAG configuration. First, we pre-calculated all cell embeddings offline to construct the CellDB, ensuring efficient dense retrieval during online inference. Second, because probabilistic QA fundamentally requires aggregating broad statistical evidence rather than locating a single factual ``needle in a haystack,'' we increased the \texttt{top-k} cell retrieval limit by $10\times$ over the default setting to maximize the evidence provided to the LLM. To ensure strict experimental fairness, TableRAG was evaluated on the exact same hardware configuration as Auto-BN and all other baselines. 

\subsection{LLM + Premise (Retrieval-Augmented Neural)}
To evaluate whether symbolic rules can improve LLM-based reasoning, we follow the CLADDER~\citep{jin2023cladder} strategy of retrieving premise statements and prompting the LLM to chain them in multiple steps (CoT). Given a question, the LLM first identifies target and evidence variables. We then retrieve up to 40 relevant premises from a pre-generated store, using either BM25, dense vector search, or a hybrid of both (see Appendix~\ref{appendix:premisestore}). These premises, alongside the question and table metadata, are passed to the LLM to calculate the answer. This setup tests the model’s ability to simulate multi-hop probabilistic derivation from one-hop rules.

Premise retrieval is implemented via LlamaIndex~\citep{Liu_LlamaIndex_2022} using Sentence-BERT~\citep{reimers-2019-sentence-bert} and BM25. We retrieve up to 40 premises per query.

\subsection{LLM + Premise + Insights (Focused Retrieval)}
\label{subsec:llm_premise_insights}

We further enhance the LLM+Premise approach by introducing a focused retrieval mechanism that incorporates only the most informative conditional patterns extracted from the Bayesian network \(\mathcal{B}\), termed \emph{insights} (as defined in \S\ref{subsec:artifacts}). Unlike unfiltered Markov Blanket contexts, which may contain redundant or weak signals compared with the base distribution, this method identifies and prioritizes high-saliency context shifts.

As formally defined in the main text (Eq.~\ref{eq:saliency}), the Insight Saliency quantifies the extent to which a context \(c\) shifts the belief distribution over a variable \(X\). For each variable, we compute this saliency score across candidate contexts generated from its Markov Blanket (MB). To filter for informative insights programmatically, we apply a set of criteria via the \textsc{IsInsight} function (Algorithm~\ref{alg:insight_extraction}): an insight must (i) exceed a minimum saliency threshold (\emph{significance threshold}, empirically set to 0.5), (ii) avoid flat distributions (\emph{dispersion threshold}, set to 0.08 based on the standard deviation), and (iii) introduce new high-probability modes relative to the base marginal \(P_X\) (\emph{new peaks}). The top-\(k\) scoring contexts that meet these conditions are retained per node.

To mitigate the combinatorial explosion inherent in large Markov Blankets, we generate conditioning contexts in a staged manner: starting with single-variable contexts and incrementally expanding to include multiple variables. This prioritized enumeration favors simpler contexts first. To ensure tractability, we implement early stopping based on two criteria: (i) a maximum number of context candidates evaluated, and (ii) a target number of valid insights extracted per node. This strategy enables efficient operation on complex, real-world Bayesian networks without exhaustive search.

By restricting the LLM input to this curated set of salient insights rather than an exhaustive collection of premises, the model is able to focus on high-precision, distributionally impactful evidence. This design improves interpretability and generalization, especially in multi-hop probabilistic reasoning scenarios. The full procedure is detailed in Algorithm~\ref{alg:insight_extraction}.

\begin{algorithm}[tb]
\caption{Insight Extraction from Bayesian Network}
\label{alg:insight_extraction}
\textbf{Input}: Bayesian Network $\mathcal{B}$, thresholds $\theta$ \\
\textbf{Output}: Top-$k$ insights $\mathcal{I}$
\begin{algorithmic}[1]
  \State Initialize $\mathcal{I} \gets \emptyset$
  \For{each variable $X \in \mathcal{B}$}
    \State $P_X \gets \text{InferWithoutContext}(\mathcal{B}, X)$
    \For{$c \in \text{GenerateContexts}(\text{MB}(X), \theta)$}
      \State $P_{X|c} \gets \text{InferWithContext}(\mathcal{B}, X, c)$
      \State $s \gets \text{Saliency}(X, c)$ \hfill \textit{// Eq.~\ref{eq:saliency}}
      \State valid $\gets \text{IsInsight}(P_{X|c}, P_X, s, \theta)$
      \If{valid}
        \State Add $(c, P_{X|c}, s)$ to top-$k$ heap $H$
      \EndIf
    \EndFor
    \State $\mathcal{I}[X] \gets$ top-$k$ elements from $H$
  \EndFor
  \State \Return $\mathcal{I}$
\end{algorithmic}
\end{algorithm}

\subsection{ProbLog-Based Symbolic Baseline}
Following QUITE~\citep{schrader-etal-2024-quite}, we attempted to encode retrieved premises into ProbLog~\citep{de2007problog} programs and evaluate queries symbolically. Conceptually, this baseline is similar to Auto-BN in that both aim to compute calibrated probabilities using a structured probabilistic engine, rather than relying on LLMs to approximate probabilistic reasoning. While theoretically appealing, this method relies on the LLM to perfectly translate each premise needed into ProbLog queries. In QUITE, this led to a failure rate exceeding 90\% even under simplified WEP settings with only ~40 premises. Due to the high brittleness and low success rate of this approach in \benchmarkname, we exclude it from final comparisons.

\subsection{Model Details}
Three types of LLM are used in our experiments: GPT-4o \cite{hurst2024gpt} (unknown size), Llama-3.3-70B~\citep{grattafiori2024llama3,meta2024llama33} (70 billion parameters), and Mixtral-8x22B~\citep{mistral2024mixtral8x22b} (141 billion total and 39 billion active parameters), a sparse mixture-of-experts model~\citep{jiang2024mixtral}, via API.
During benchmark curation,  GPT-4o is used to generate natural language text. All baselines involving LLMs use consistent prompting templates, with GPT-4o serving as the primary model during development. LLaMA-3.3-70B and Mixtral-8x22B are evaluated using the same prompts without additional fine-tuning.

The Bayesian Networks training for Auto-BN uses PGMPY\cite{Ankan2024} library on an r5.4xlarge AWS instance (8 vCPUs, 128 GB RAM, Amazon Linux 2023 system); training time is reported in Appendix~\ref{subsec:running_time}

\section{Additional Experimental Results}
\label{sec-addiexp}

\subsection{Performance by Question Type}

Figure~\ref{fig:metrics_question_type} presents the performance of various baselines across different question types. For each type, we report three metrics: $\text{Acc}_{0.02}$ (↑), $\text{Acc}_{0.05}$ (↑), and MAE (↓), capturing both accuracy and error magnitude.

Across all question types, Auto-BN consistently outperforms the other baselines, achieving the highest accuracy and lowest error. Even in the more challenging \textit{Natural} questions, Auto-BN maintains superior performance, highlighting its robustness under linguistic variability. The four variants stress different pipeline steps. \textit{Natural} questions replace numeric conditions with semantic descriptors (e.g., ``high temperature''), which are ambiguous when mapped to table values and BN bin boundaries. \textit{Certain} questions state a single value that must be mapped to its containing BN state (e.g., ``exactly 7 goals'' $\rightarrow$ bin $(4, 10]$); few or no records match such a point conjunction exactly, so record-counting methods such as NL2SQL break down (Figure~\ref{fig:error_nl2sql}). \textit{Shifted} questions use an alternative valid range (e.g., ``between 4 and 7'' when the BN state is $(4, 10]$), which the model must recognize as a sub-range of an existing state. LLM mathematical reasoning has been shown to degrade when numerical values move outside typical ranges \citep{shrestha2025mathematical}, and the narrower range again reduces the number of matching records.

The LLM+Table baseline performs relatively well on the \textit{Certain} question type. This is likely because questions specifying a single numeric value can be more directly matched against sampled rows, allowing the LLM to approximate probabilities more reliably than with binned numeric states.

In contrast, the NL2SQL baseline degrades as the question type becomes more natural or paraphrased, since such formulations are harder to translate into valid and expressive SQL clauses. This confirms the brittleness of deterministic symbolic pipelines under linguistic variation.

Adding Insights consistently improves the LLM+Premise baselines across all question types. This highlights the importance of retrieving high-impact, distributional evidence rather than relying solely on shallow one-hop premises, as it leads to better reasoning fidelity and calibration.

\subsection{Performance by Probabilistic Query Type}

\begin{table*}[!htbp]
\centering
\small
\renewcommand{\arraystretch}{1.0}
\setlength{\tabcolsep}{2pt}
\resizebox{\textwidth}{!}{%
\begin{tabular}{llcccccccccc}
\toprule
\textbf{Type} & \textbf{Method} 
& \multicolumn{2}{c}{\shortstack{\textbf{Forward}\\\textbf{Conditional}}} 
& \multicolumn{2}{c}{\shortstack{\textbf{Diagnostic}}} 
& \multicolumn{2}{c}{\shortstack{\textbf{Interdependent}\\\textbf{Evidence}}} 
& \multicolumn{2}{c}{\textbf{Mixed}} 
& \multicolumn{2}{c}{\shortstack{\textbf{Mixed+} \\ \textbf{Redundant}}} \\
\cmidrule(l){3-4} \cmidrule(l){5-6} \cmidrule(l){7-8} \cmidrule(l){9-10} \cmidrule(l){11-12}
& & Acc$_{0.02}$ & RMSE & Acc$_{0.02}$ & RMSE & Acc$_{0.02}$ & RMSE & Acc$_{0.02}$ & RMSE & Acc$_{0.02}$ & RMSE \\
\specialrule{1pt}{1pt}{1pt}
\addlinespace[0.3em]

\multirow{4}{*}{Factual}
  & LLM + Table & 9.6 & 0.234 & 12.8 & 0.212 & 10.2 & 0.229 & 13.9 & 0.248 & 11.5 & 0.230 \\
  & TableRAG & 6.6 & 0.304 & 5.5 & 0.297 & 10.2 & 0.258 & 8.4 & 0.319 & 5.5 & 0.320 \\
  & Reasoning (GPT-5.4) & 11.0 & 0.268 & 12.1 & 0.276 & 12.2 & 0.281 & 11.4 & 0.304 & 10.0 & 0.287 \\
  & NL2SQL & 12.6 & 0.238 & 16.6 & 0.242 & 13.1 & 0.261 & 16.4 & 0.285 & 8.9 & 0.249 \\

\cmidrule{1-12}

\multirow{3}{*}{\shortstack{Prob. \\ Premises}}
  & LLM+Premise(Vector) & 10.3 & 0.207 & 8.8 & 0.237 & 7.8 & 0.277 & 8.0 & 0.255 & 6.1 & 0.228 \\
  & LLM+Premise(BM25) & 9.0 & 0.218 & 9.0 & 0.261 & 5.7 & 0.259 & 10.5 & 0.277 & 10.0 & 0.231 \\
  & LLM+Premise(Hybrid) & 11.0 & 0.195 & 10.7 & 0.235 & 7.6 & 0.241 & 11.7 & 0.252 & 10.6 & 0.198 \\

\cmidrule{1-12}

\multirow{3}{*}{\shortstack{Prob. \\ Insights}}
  & LLM+Premise(Vector)+Insights & 17.2 & 0.179 & 19.8 & 0.211 & \textbf{39.0} & 0.202 & 19.5 & 0.218 & 24.8 & 0.177 \\
  & LLM+Premise(BM25)+Insights & 13.9 & 0.193 & 19.3 & 0.217 & 38.6 & 0.182 & 17.6 & 0.239 & 25.0 & 0.186 \\
  & LLM+Premise(Hybrid)+Insights & 15.0 & 0.182 & 21.3 & 0.218 & 37.2 & 0.181 & 20.3 & 0.232 & 24.1 & \textbf{0.167} \\

\cmidrule{1-12}

\multirow{1}{*}{}
  & Auto-BN & \textbf{28.7} & \textbf{0.151} & \textbf{31.1} & \textbf{0.151} & 32.9 & \textbf{0.152} & \textbf{22.3} & \textbf{0.208} & \textbf{25.1} & 0.177 \\

\bottomrule
\end{tabular}%
}
\caption{Performance across probabilistic query types using GPT-4o. Each method is grouped by baseline type and evaluated on $\text{Acc}_{0.02}$ (↑) and RMSE (↓) per reasoning category.}
\label{table:reasoning-type-gpt4o}
\end{table*}

Table~\ref{table:reasoning-type-gpt4o} reports zero-shot GPT-4o performance ($\text{Acc}_{0.02}$ and RMSE) across the five probabilistic reasoning types, averaged over all linguistic question variants.

Overall, Auto-BN achieves the strongest performance in 8 of 10 metrics, demonstrating exceptional robustness across \textit{Forward Conditional}, \textit{Diagnostic}, and \textit{Mixed} queries. This proves that exact mathematical inference over an induced graph consistently outperforms purely generative probability estimation. Auto-BN is only outperformed in \textit{Interdependent Evidence} ($\text{Acc}_{0.02}$) and \textit{Mixed+Redundant} (RMSE) by insight-augmented baselines. This stems from an inherent information asymmetry: \textit{Insights} are derived directly from the human-verified Gold BNs. In highly complex or redundant contexts, these oracle-calibrated signals help the LLM filter noise more effectively than Auto-BN's fully automated, purely observational graph.

Among retrieval-based methods, \textit{Insights} consistently outperform raw premises by distilling massive premise spaces into compact, high-impact conditional shifts. Conversely, factual (NL2SQL, LLM+Table) and premise-only baselines fail uniformly on composite queries (\textit{Mixed} types). Such queries require chaining heterogeneous statistical cues and ignoring distractors—capabilities that deterministic SQL or shallow text retrieval fundamentally lack.

\begin{table*}[!htbp]
\centering
\resizebox{\linewidth}{!}{%
\begin{tabular}{lcccccc}
\toprule
\textbf{Method} & \textbf{Premises} & \textbf{\% Acc\textsubscript{0.02}} $\uparrow$ & \textbf{\% Acc\textsubscript{0.05}} $\uparrow$ & \textbf{MAE} $\downarrow$ & \textbf{RMSE} $\downarrow$ & \textbf{\% Error Rate} $\downarrow$ \\
\midrule

\multirow{3}{*}{LLM+Premise(Vector)} & 1-hop & 8.2 & 24.4 & 0.162 & 0.226 & 0.0 \\
 & 2-hop & 9.2 & 24.1 & 0.165 & 0.231 & 0.0 \\
 & \textbf{Improvement} & \textbf{0.9} & \textbf{-0.3} & \textbf{0.003} & \textbf{0.005} & \textbf{0.0} \\
\midrule

\multirow{3}{*}{LLM+Premise(Vector)+Insights} & 1-hop & 25.7 & 42.7 & 0.104 & 0.173 & 0.0 \\
 & 2-hop & 24.4 & 40.2 & 0.113 & 0.184 & 0.0 \\
 & \textbf{Improvement} & \textbf{-1.3} & \textbf{-2.6} & \textbf{0.008} & \textbf{0.011} & \textbf{0.0} \\
\bottomrule
\end{tabular}
}
\caption{1-Hop vs 2-Hop Premises Ablation Study on \textit{cookbook nutrition} Dataset}
\label{tab:1hop_2hop_comparison}
\end{table*}

\subsection{Per-Dataset Results}
Table~\ref{tab:per-dataset} reports MAE by dataset for the main baselines. Auto-BN achieves the lowest or tied-lowest MAE on 8 of 10 datasets. The two exceptions, \textit{coinmarketcap} and \textit{college completion}, favor LLM+Premise(H)+Ins (0.100 vs.\ 0.117; 0.142 vs.\ 0.184). These datasets may be harder for automatic structure learning or may benefit more from the Gold-BN-derived premises and insights that Auto-BN does not use. Auto-BN's MAE also varies more across datasets (0.038--0.184) than that of the premise- and insight-based baselines. Adding insights to retrieved premises improves MAE on all 10 datasets.

\begin{table*}[h]
\centering
\small
\begin{tabular}{l r r r r r r}
\toprule
\textbf{Dataset} & \textbf{LLM+Table} & \textbf{Reasoning} & \textbf{NL2SQL} & \textbf{LLM+Premise(H)} & \textbf{+Insights} & \textbf{Auto-BN} \\
\midrule
airlines              & 0.175 & 0.212 & 0.200 & 0.140 & 0.084 & \textbf{0.038} \\
bike stations weather & 0.154 & 0.218 & 0.150 & 0.165 & 0.148 & \textbf{0.099} \\
coinmarketcap         & 0.235 & 0.245 & 0.237 & 0.133 & \textbf{0.100} & 0.117 \\
college completion    & 0.164 & 0.201 & 0.198 & 0.172 & \textbf{0.142} & 0.184 \\
cookbook nutrition    & 0.167 & 0.170 & 0.227 & 0.156 & 0.104 & \textbf{0.102} \\
hockey scoring        & 0.188 & 0.190 & 0.252 & 0.214 & 0.171 & \textbf{0.120} \\
movies                & 0.151 & 0.164 & 0.145 & 0.131 & 0.116 & \textbf{0.090} \\
basketball players    & 0.169 & 0.226 & 0.161 & 0.213 & 0.171 & \textbf{0.123} \\
sales in weather      & 0.164 & 0.209 & 0.167 & 0.166 & 0.144 & \textbf{0.101} \\
twitter               & 0.145 & 0.152 & 0.161 & 0.139 & \textbf{0.090} & \textbf{0.090} \\
\bottomrule
\end{tabular}
\caption{Per-dataset MAE ($\downarrow$). All methods use GPT-4o except Reasoning (GPT-5.4). LLM+Premise(H) uses Hybrid retrieval; +Insights adds insights to it. Best per row in bold.}
\label{tab:per-dataset}
\end{table*}

\subsection{Extending the One-Hop Premise Definition}

While most prior benchmarks operate on one-hop premises (grounded conditional probability statements of the form $P(Y | X)$ representing direct parent–child dependencies in the underlying conditional graph), we examine whether incorporating multi-hop premises (specifically two-hops) can improve performance by capturing richer contextual dependencies. However, extending to multi-hop contexts increases combinatorial complexity and may introduce noisy or redundant information. To focus the analysis, we restrict our study to two-hop premises and evaluate their impact on the \textit{cookbook nutrition} dataset. We construct two-hop premise representations by extracting all valid, unique two-step conditional dependencies and combining them with existing one-hop premises to form an augmented premise vector store. Both the LLM+Premise(Vector) and its Insight-augmented variant are then evaluated under this expanded evidence setup.

Table~\ref{tab:1hop_2hop_comparison} summarizes the ablation results. For the LLM+Premise(Vector) variant, introducing 2-hop premises yields a minor gain in $\text{Acc}_{0.02}$ (+0.9), but slightly degrades $\text{Acc}_{0.05}$, MAE, and RMSE, indicating potential trade-offs between local accuracy and global calibration.

Interestingly, the Insight-augmented method performs better when restricted to 1-hop premises, with performance declining across all metrics when 2-hop premises are included in this dataset. This suggests that the insight extraction mechanism, designed to identify salient signals, may be adversely affected by the increased noise and reduced contextual relevance of deeper dependency chains.

These preliminary results suggest that, although multi-hop premises can provide longer reasoning chains and broader contextual coverage, their utility appears to depend strongly on precision and relevance. In the absence of strong filtering mechanisms, deeper contexts may introduce noise that reduces the quality of retrieved evidence and hinders downstream reasoning. Further evaluation across diverse datasets is needed to determine when and how multi-hop contexts can be incorporated effectively.

\subsection{Running Time Analysis}\label{subsec:running_time}

\begin{table*}[!htbp]
\centering
\renewcommand{\arraystretch}{0.90}
\small
\begin{tabular}{lrrrrr}
\toprule
\textbf{Dataset} & \textbf{Training (s)} & \textbf{Inference (s)} & \textbf{Total (s)} & \textbf{Inference/Q (s)} & \textbf{Total/Q (s)} \\
\midrule
airlines            &    242 &    935 &  1,177 &  1.49 &  1.87 \\
basketball players  &    852 &  1,022 &  1,874 &  1.63 &  2.98 \\
bike stations weather & 1,057 & 1,084 &  2,141 &  1.73 &  3.41 \\
coinmarketcap       &  5,950 &  1,067 &  7,017 &  1.70 & 11.17 \\
college completion  &    218 &    981 &  1,199 &  1.56 &  1.91 \\
cookbook nutrition  &    287 &  1,038 &  1,325 &  1.65 &  2.11 \\
hockey scoring      &    632 &  1,171 &  1,803 &  1.87 &  2.87 \\
movies              &     94 &    975 &  1,069 &  1.55 &  1.70 \\
sales in weather    &  1,176 &  2,569 &  3,745 &  4.09&  5.96 \\
twitter             &    100 &  1,065 &  1,165 &  1.70 &  1.86 \\
\midrule
\textbf{Mean}       &  1,061 &  1,191 &  2,252 &  1.90 &  3.58 \\
\bottomrule
\end{tabular}
\caption{Training and inference time statistics for Auto-BN pipeline using GPT-4o across datasets (157 questions $\times$ 4 question types = 628 queries per dataset). Training is a one-time offline cost. Inference/Q and Total/Q denote mean seconds per query.}
\label{tab:running_time}
\end{table*}

Auto-BN comprises an offline training phase (dependency graph discovery and BN parameter learning) and an online inference phase (robust query translation and exact probabilistic execution). Table~\ref{tab:running_time} summarizes the computational footprint of both phases across the benchmark.

\paragraph{Offline Induction Cost.}
As shown in Table~\ref{tab:running_time}, the one-time offline induction scales naturally with dataset complexity. It ranges from 94 seconds (\textit{movies}) to 5,950 seconds (\textit{coinmarketcap}), the latter requiring extensive structural search over numerous attributes. Crucially, this induction is a CPU-only process executed exactly once per table. When amortized over the lifespan of a deployed database, this offline cost becomes negligible.

\paragraph{Online Inference Latency.}
During the online phase, inference latency averages a highly efficient 1.90 seconds per query across the benchmark. To maintain this low overhead despite the rigorous Multi-Path Cross-Consistency Voting (\S\ref{subsec:robust_translation}), Auto-BN executes the $N{=}3$ LLM parsing calls concurrently, bounding wall-clock time strictly by the slowest individual API response. The only notable outlier is \textit{sales in weather} (4.09 s/q), where exceptionally high BN edge density inherently increases the computational complexity of exact variable elimination. Overall, this 1.90-second latency is highly competitive with LLM+RAG methods (e.g. TableRAG ${\sim}4.51$ s/q) while delivering the critical advantage of mathematically grounded, hallucination-free probabilistic guarantees.

\subsection{Ablation Study: Robust Query Translation}\label{subsec:error_analysis}

\begin{figure*}[!ht]
   \centering
   \includegraphics[width=\linewidth]{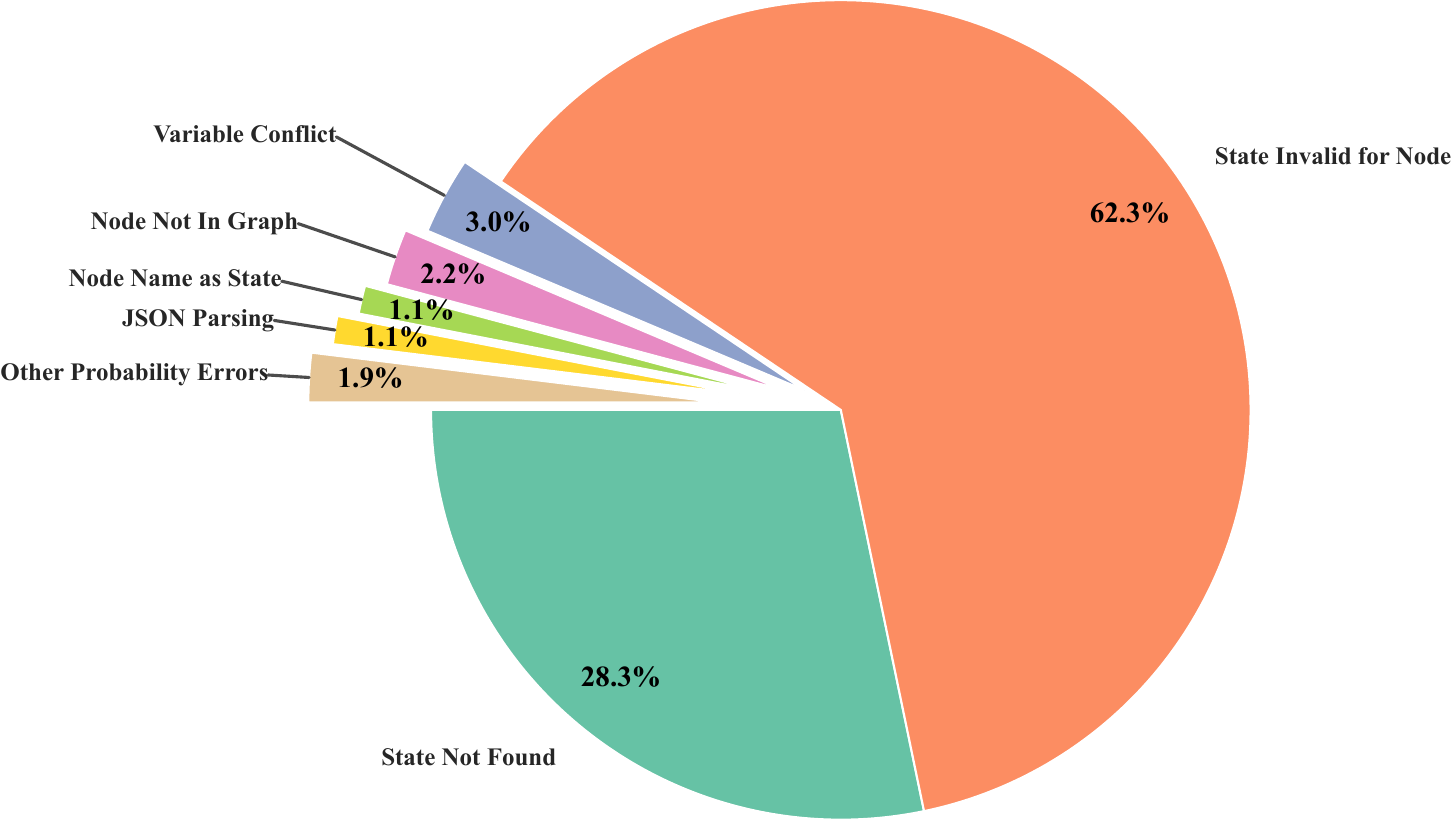}
   \caption{Distribution of parsing errors in the Base Auto-BN ablation configuration (single-pass parse, no correction mechanisms). Percentages are averaged across all datasets. Auto-BN resolved these errors.}
   \label{fig:error_analysis}
\end{figure*}

A critical bottleneck in neuro-symbolic pipelines is the inherent brittleness of LLM-based semantic parsing. To demonstrate how Auto-BN overcomes this limitation, we ablate our robust query translation and multi-level relaxation mechanisms (\S\ref{subsec:robust_translation}, \S\ref{subsec:multi_level_relaxation}) and analyze the resulting failure modes. We define a \textbf{Base Auto-BN} configuration that relies exclusively on a single zero-shot parse, completely removing multi-path voting, execution-guided refinement, and all inference-level fallbacks. Figure~\ref{fig:error_analysis} illustrates the exact error distribution for this uncorrected baseline across all benchmark datasets.

The dominant failure categories are: \textit{State Invalid for Node} (62.3\%), where the LLM predicts a state that does not belong to the queried node; \textit{State Not Found} (28.3\%), where the predicted state does not exist anywhere in the BN vocabulary; \textit{Variable Conflict} (3.0\%), where the same variable appears as both target and evidence; \textit{Node Not in Graph} (2.2\%); \textit{Node Name as State} (1.1\%); \textit{JSON Parsing errors} (1.1\%); and \textit{Other Inference Errors} (1.9\%). 

Over 90\% of these baseline failures stem from node--state vocabulary mismatches, confirming that without structural safeguards, LLMs systematically hallucinate states outside the BN schema. These findings directly motivate our proposed pipeline: (i)~multi-path cross-consistency voting to surface schema-invalid parses early, (ii)~execution-guided iterative refinement to self-correct mismatches, and (iii)~semantic relaxation fallbacks to recover the nearest valid node or state when strict parsing fails.

Table~\ref{tab:ablation_robust} quantifies the impact of these mechanisms across all three LLMs on the full \benchmarkname benchmark, with an \textit{Improvement} row showing the absolute delta (Auto-BN $-$ Base Auto-BN) per LLM. By successfully insulating the exact probabilistic inference engine from the stochastic brittleness of LLM generation, the full Auto-BN framework reduces the error rate from as high as 20.8\% down to near-zero, while simultaneously drives consistent gains in strict accuracy ($\text{Acc}_{0.02}$ and $\text{Acc}_{0.05}$) and maintaining stable MAE and RMSE.

\begin{table*}[!ht]
\centering
\small
\begin{tabular}{l l r r r r r}
\toprule
\textbf{LLM} & \textbf{Method} & \textbf{\% Acc\textsubscript{0.02}} $\uparrow$ & \textbf{\% Acc\textsubscript{0.05}} $\uparrow$ & \textbf{MAE} $\downarrow$ & \textbf{RMSE} $\downarrow$ & \textbf{\% Err} $\downarrow$ \\
\midrule
\multirow{3}{*}{GPT-4o}
  & Base Auto-BN        & $26.4 \pm 14.6$ & $45.3 \pm 15.3$ & $0.107 \pm 0.040$ & $0.168 \pm 0.055$ & $5.6 \pm 4.6$ \\
  & Auto-BN             & $28.0 \pm 15.5$ & $46.7 \pm 16.2$ & $0.107 \pm 0.043$ & $0.169 \pm 0.058$ & $0.0 \pm 0.0$ \\
  & Improvement         & $+1.6$ & $+1.4$ & $0.000$ & $+0.001$ & $-5.6$ \\
\midrule
\multirow{3}{*}{Llama-3.3-70B}
  & Base Auto-BN        & $23.6 \pm 12.8$ & $41.6 \pm 14.4$ & $0.111 \pm 0.036$ & $0.169 \pm 0.047$ & $14.8 \pm 10.6$ \\
  & Auto-BN             & $25.4 \pm 13.6$ & $44.8 \pm 15.0$ & $0.109 \pm 0.040$ & $0.168 \pm 0.053$ & $0.0 \pm 0.0$ \\
  & Improvement         & $+1.8$ & $+3.2$ & $-0.002$ & $-0.001$ & $-14.8$ \\
\midrule
\multirow{3}{*}{Mixtral-8x22B}
  & Base Auto-BN        & $22.7 \pm 13.9$ & $40.4 \pm 15.1$ & $0.112 \pm 0.033$ & $0.167 \pm 0.043$ & $20.8 \pm 12.0$ \\
  & Auto-BN             & $26.1 \pm 15.4$ & $44.6 \pm 16.1$ & $0.112 \pm 0.046$ & $0.174 \pm 0.062$ & $0.0 \pm 0.2$ \\
  & Improvement         & $+3.4$ & $+4.2$ & $0.000$ & $+0.007$ & $-20.8$ \\
\bottomrule
\end{tabular}
\caption{Ablation of robust query translation on LUCARIO. Auto-BN (this work) adds multi-path voting, execution-guided refinement, and inference-level fallbacks over Base Auto-BN (single-pass parse, no correction). \textit{Improvement} rows show absolute deltas (Auto-BN $-$ Base Auto-BN); negative values for MAE/RMSE/Err indicate improvement.}
\label{tab:ablation_robust}
\end{table*}

\subsection{Case Study} \label{sec_case_study}

\begin{figure}
    \centering
    \includegraphics[width=1.0\linewidth]{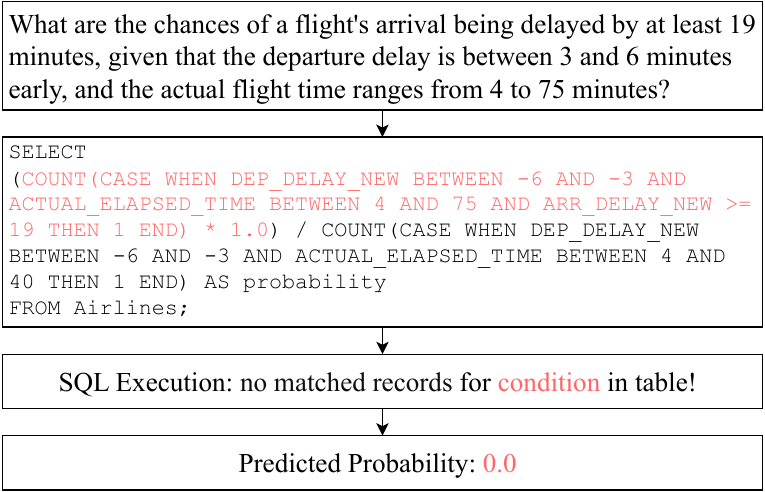}
    \caption{Case Study for Factual Method on Prob. QA}
    \label{fig:error_nl2sql}
\end{figure}

Why factual retrieval based method doesn't work well on prob table QA?
Figure \ref{fig:error_nl2sql} shows one frequent failure that NL2SQL baseline made. Even if NL2SQL model successfully translates the probabilistic question into valid SQL query, the context provided in the question might not find exact match records in the table. That causes NL2SQL baseline frequently output 0.0 (no records) or 1.0 (bias in retrieved few records).

\section{Gold BN vs Induced BN}\label{appendix:gold-bn}

In our main evaluation, the proposed Auto-BN framework operates in a strict, fully automated setting—inducing its Bayesian Network directly from raw tabular data without human intervention. To establish the upper bound of this benchmark and isolate the impact of automatic structure induction, we compare the standard Auto-BN against an ablation using the human-verified Gold BN. 

This analysis serves two purposes: First (\S~\ref{subsec:gold-bn-qa}), it establishes one type of benchmark's performance ceiling when the underlying mathematical structure is perfectly specified. Second (\S~\ref{subsec:gold-bn-structure}), it quantifies the structural differences between the induced and Gold BNs, confirming that Auto-BN operates without information leakage and validating its strong performance in the harder, fully automated regime.

\subsection{QA Performance: Benchmark Upper Bound}\label{subsec:gold-bn-qa}

To establish the upper bound, we introduce \textbf{Auto-BN (Gold)}. This ablation uses the exact same execution-guided translation, voting, and multi-level fallback pipeline described in \S~\ref{sec-method}, but replaces the automatically induced BN with the benchmark's human-verified ground-truth BN.

\begin{table*}[!htbp]
\centering
\small
\setlength{\tabcolsep}{4pt}
\begin{tabular}{llcccc}
\toprule
\textbf{Method} & \textbf{LLM} & \textbf{\% Acc\textsubscript{0.02}} $\uparrow$ & \textbf{\% Acc\textsubscript{0.05}} $\uparrow$ & \textbf{MAE} $\downarrow$ & \textbf{RMSE} $\downarrow$ \\
\midrule
\multirow{2}{*}{Auto-BN (Gold)} & GPT-4o & \textbf{77.9} & \textbf{83.8} & \textbf{0.034} & \textbf{0.085} \\
 & Llama-3.3-70B & 66.4 & 75.4 & 0.049 & 0.108 \\
 & Mixtral-8x22B & 67.5 & 76.4 & 0.050 & 0.111 \\
\midrule
\multirow{3}{*}{Auto-BN (Induced)} & GPT-4o & 28.0 & 46.7 & 0.107 & 0.169 \\
 & Llama-3.3-70B & 25.4 & 44.8 & 0.109 & 0.168 \\
 & Mixtral-8x22B & 26.3 & 45.4 & 0.109 & 0.168 \\
\bottomrule
\end{tabular}
\caption{QA performance ablation isolating the impact of graph structure quality. \textit{Auto-BN (Gold)} represents the upper bound when provided an expert-curated schema. Metrics are averaged over 6,280 total questions.}
\label{tab:gold-bn-qa}
\end{table*}

When equipped with the human-verified BN, Auto-BN (Gold) with GPT-4o achieves an outstanding $\text{Acc}_{0.05}$ of 83.8\% and a near-zero MAE of 0.034. This establishes a strong upper bound for the benchmark, demonstrating that the proposed query translation pipeline is highly effective when the underlying probabilistic structure is accurately specified.

By contrast, the standard Auto-BN (Induced) operates at an MAE of roughly 0.107. Since the translation prompts and fallback mechanisms are strictly identical in both settings, this performance gap reflects the inherent difficulty of purely observational structure induction. However, achieving an MAE of 0.107 in a purely automated, zero-shot setting remains a highly competitive result, demonstrating that Auto-BN provides grounded, reasonable probabilistic estimates even without expert domain knowledge.

\subsection{Structural Fidelity of Induced BNs}\label{subsec:gold-bn-structure}

To structurally explain the performance gap observed above, we evaluate how well the data-driven PC algorithm recovers the Gold graph structure. We compare the edge sets of the Induced BN against the Gold BN across all 10 datasets, computing precision (P), recall (R), and F$_1$ scores in two settings: (1)~\textbf{Directed}: an induced edge $u \rightarrow v$ is a true positive only if the gold graph contains the exact same directed dependency; and (2)~\textbf{Undirected}: an induced edge between $u$ and $v$ is a true positive if the gold graph connects them in either direction.

\begin{table}[!ht]
\centering
\small
\setlength{\tabcolsep}{5pt}
\begin{tabular}{l|ccc|ccc}
\toprule
 & \multicolumn{3}{c|}{\textbf{Directed}} & \multicolumn{3}{c}{\textbf{Undirected}} \\
 & \textbf{P} & \textbf{R} & \textbf{F$_1$} & \textbf{P} & \textbf{R} & \textbf{F$_1$} \\
\midrule
Mean & 0.165 & 0.156 & 0.159 & 0.582 & 0.563 & 0.569 \\
Std  & 0.105 & 0.105 & 0.103 & 0.236 & 0.266 & 0.248 \\
\bottomrule
\end{tabular}
\caption{Edge-level structural fidelity of the automatically induced BN vs. the human-verified Gold BN.}
\label{tab:bn-structure}
\end{table}

\paragraph{Structural Divergence and Robustness.} 
The undirected F$_1$ score averages 0.569, indicating that the automatic induction algorithm successfully identifies over half of the underlying statistical associations between variables. This is a robust recovery rate, particularly considering that there are often multiple valid ways to represent similar hidden relationships among columns (e.g., modeling a direct 1-hop dependency via an intermediate 2-hop path). However, the directed F$_1$ drops severely to 0.159. This profound structural divergence mathematically confirms that the automatically induced BN is distinctly different from the human-verified Gold BN, definitively proving there is no leakage of expert domain knowledge into our automated pipeline. It also validates that Auto-BN and premise-based baselines (e.g., LLM+Premise) operate under fundamentally different information conditions.

While Auto-BN can recover from semantic vocabulary mismatches, it cannot mathematically compensate for systematically inverted dependency directions in the induced graph. In light of this significant structural handicap, Auto-BN's ability to still achieve an MAE of 0.107 underscores the exceptional robustness of the proposed framework, proving its practical viability and strong performance for purely data-driven tabular reasoning.

\section{Auto-BN-Tool (Agentic BN Inference)}\label{appendix:autobn_tool}

To facilitate the integration of our proposed system into modern agentic workflows, we introduce \textbf{Auto-BN-Tool}, a tool-calling extension of the base Auto-BN framework. By wrapping the exact probabilistic inference engine into an executable tool, Auto-BN-Tool allows the LLM to operate within a multi-turn agent loop. The model can iteratively invoke the tool, analyze execution results or error messages, and synthesize a final answer from multiple sub-queries. This formulation demonstrates the flexibility of our modular design, showing that Auto-BN can seamlessly serve as a foundational probabilistic reasoning engine for autonomous LLM agents.

In practice, at each turn, the LLM is prompted with the natural language question alongside the derived BN metadata (i.e., node and state vocabularies) and the history of prior tool executions. The LLM is explicitly instructed to decompose complex questions into manageable sub-queries. It then either emits a \texttt{<tool\_call>} block to compute a specific probability on the BN or outputs the final calculated answer. The agentic loop terminates when the LLM produces a final answer or reaches a predefined budget ($T_\text{max}{=}5$ turns, with a maximum of $C_\text{max}{=}3$ tool calls per run).

This iterative formulation offers two primary advantages. First, it enables the LLM to handle complex, multi-step reasoning by issuing sequential sub-queries (e.g., verifying a marginal distribution before querying a conditional one). Second, it acts as an intrinsic error-correction mechanism. If the LLM hallucinates an invalid node or state, the tool returns an execution error. The agent can then observe this feedback, consult the provided BN vocabulary, and self-correct its semantic mapping in the subsequent turn.

\begin{table*}[!htp]
\centering
\resizebox{\linewidth}{!}{%
\small
\begin{tabular}{l c c c c c}
\toprule
\textbf{Method} & \textbf{\% Acc\textsubscript{0.02}} $\uparrow$ & \textbf{\% Acc\textsubscript{0.05}} $\uparrow$ & \textbf{MAE} $\downarrow$ & \textbf{RMSE} $\downarrow$ & \textbf{\% Error} $\downarrow$ \\
\midrule
Auto-BN & $\mathbf{28.0 \pm 15.5}$ & $46.7 \pm 16.2$ & $0.107 \pm 0.043$ & $0.169 \pm 0.058$ & $\mathbf{0.0 \pm 0.0}$ \\
Auto-BN-tool & $27.8 \pm 15.7$ & $\mathbf{47.6 \pm 16.5}$ & $\mathbf{0.105 \pm 0.043}$ & $\mathbf{0.166 \pm 0.057}$ & $0.1 \pm 0.4$ \\
\bottomrule
\end{tabular}
}
\caption{Performance of the agentic Auto-BN-Tool framework on the Lucario benchmark.}
\label{table:autobn-tool-results}
\end{table*}

We evaluate Auto-BN-Tool using GPT-4o as the backbone LLM. As detailed in Table~\ref{table:autobn-tool-results}, the agentic framework delivers strong overall performance, achieving an $Acc_{0.05}$ of 47.6\%, a MAE of 0.105, and an RMSE of 0.166. Furthermore, its strict accuracy (27.8\% $Acc_{0.02}$) remains highly competitive with the base Auto-BN pipeline (28.0\%) while successfully maintaining a near-zero error rate (0.1\%). Ultimately, these results demonstrate that integrating Auto-BN into an autonomous agent loop not only preserves the rigorous, hallucination-free grounding of exact inference, but also maintains system robustness and calibration through multi-turn self-correction and compositional query decomposition.

\section{Prompt Templates}
\label{appendix:prompts}

This appendix provides the exact prompt templates used in our experiments. Runtime variables are denoted in lowercase within curly braces, e.g., \texttt{\{question\}}.

\subsection{Auto-BN: Semantic Parsing Prompt}
\label{app:prompt_autobn}

The following system prompt is used to translate a natural language question into a structured probabilistic query (target node, target state, evidence context) for exact BN inference. This prompt is used by \textbf{Auto-BN}.

\begin{promptbox}{Auto-BN Semantic Parsing Prompt}
You are an assistant that maps natural language questions into probabilistic components of a Bayesian Network. The Bayesian Network is built from a tabular data, where each node corresponds to a column in the original table.

A Bayesian Network consists of:\\
-- \textbf{Nodes} (random variables)\\
-- \textbf{States} (possible values each node can take)\\
-- \textbf{Directed edges} indicating conditional dependencies between variables

This network encodes a joint probability distribution P(X$_1$, X$_2$, ..., X$_n$). Each edge A $\rightarrow$ B means that the conditional probability P(B $|$ A) is directly represented.

\vspace{2pt}
\textbf{Nodes and their states:} \{nodes\}\\
\textbf{Edges} (parent $\rightarrow$ child): \{edges\}\\
\textbf{Descriptions of the nodes (from the tabular data):} \{table\_description\}

\vspace{2pt}
Given the following question, extract:\\
(1) The \textbf{target node} being queried (the variable we want the probability for)\\
(2) The \textbf{target state} or value of that variable\\
(3) Any \textbf{observed context} or evidence described in the question (as a dictionary of node $\rightarrow$ state)

\textbf{Question:} \{question\}

\vspace{2pt}
\textbf{CRITICAL INSTRUCTIONS:}\\
-- You MUST use ONLY the exact node names and state values provided above\\
-- Do NOT create new state values or modify existing ones\\
-- If the question mentions a range that does not exactly match a provided state, choose the CLOSEST available state\\
-- Double-check that your selected states exist in the provided list before responding

Respond ONLY with a JSON object in the exact format below:\\
\{\ "target\_node": "node\_name",\ "target\_state": "state\_value",\ "context": \{"node1": "state1"\}\ \}
\end{promptbox}

\subsection{Auto-BN-Tool: Agentic Loop Prompt}
\label{app:prompt_tool}

The following system prompt governs the Auto-BN-Tool agentic loop (Appendix~\ref{appendix:autobn_tool}). The LLM reasons step-by-step, issuing \texttt{<tool\_call>} blocks to invoke exact BN inference, observing results, and emitting a final answer via \texttt{<final\_answer>} tags.

\begin{promptbox}{Auto-BN-Tool Agentic System Prompt}
You are a probabilistic reasoning agent that answers questions about tabular data using a pre-learned Bayesian Network (BN).

\vspace{2pt}
=== Bayesian Network Structure ===\\
Nodes and their possible states: \{nodes\}\\
Edges (parent $\rightarrow$ child): \{edges\}\\
Column descriptions: \{table\_description\}

\vspace{2pt}
=== Tool: query\_bayesian\_network ===\\
To invoke the tool, output a JSON block wrapped in \texttt{<tool\_call>} tags:\\
\texttt{<tool\_call>}\{"target\_node": "NodeName", "target\_state": "StateValue", "context": \{"N1": "S1"\}\}\texttt{</tool\_call>}

You may make up to \{max\_tool\_calls\} tool call(s). The tool returns:\\
\{"status": "success"/"error",\ "probability": 0--1 or null,\ "fallback\_used": ...\}

\vspace{2pt}
=== Providing your final answer ===\\
Mark your final answer with: \texttt{<final\_answer>0.42</final\_answer>}\\
The value must be a number in [0, 1]. Nothing should appear after this tag.

\vspace{2pt}
=== Reasoning strategy ===\\
(1) Identify the target variable, its value, and any conditioning evidence\\
(2) \textbf{Decompose into sub-queries} only when the question explicitly requires combining independent probabilities (e.g., computing P(A) and P(B$|$C) separately), or when an initial call fails and reformulation is needed; simple conditional queries P(A$|$B) should be issued as a single tool call\\
(3) Call \texttt{query\_bayesian\_network} with the identified target and context\\
(4) If the tool returns an error, check node/state names against the schema and retry with corrected names\\
(5) Synthesize the sub-query results into a final probability estimate and emit your final answer using \texttt{<final\_answer>X</final\_answer>}
\end{promptbox}

\end{document}